\documentclass[sigconf]{acmart}
\AtBeginDocument{%
  }

\copyrightyear{2025}
\acmYear{2025}
\setcopyright{acmlicensed}

\acmConference[MM '25] {Proceedings of the 33rd ACM International Conference on Multimedia}{October 27--31, 2025}{Dublin, Ireland.}
\acmBooktitle{Proceedings of the 33rd ACM International Conference on Multimedia (MM '25), October 27--31, 2025, Dublin, Ireland}
\acmDOI{10.1145/3746027.3754740}
\acmISBN{979-8-4007-2035-2/2025/10}
\usepackage{multirow}
\usepackage{microtype}
\usepackage{wrapfig}
\usepackage{balance}

\begin{document}

\title{EmbodiedOcc++: Boosting Embodied 3D Occupancy Prediction with Plane Regularization and Uncertainty Sampler}

\author{Hao Wang}
\authornote{Equal contribution to this research.}
\orcid{0009-0006-0208-6284}
\affiliation{%
  \institution{State Key Laboratory of Multimedia Information Processing, School of Computer Science, Peking University}
  \state{Beijing}
  \country{China}
}
\email{haowang@stu.pku.edu.cn}

\author{Xiaobao Wei}
\authornotemark[1]
\orcid{0000-0003-4230-1162}
\affiliation{%
  \institution{State Key Laboratory of Multimedia Information Processing, School of Computer Science, Peking University}
  \state{Beijing}
  \country{China}
}
\email{weixiaobao23@mails.ucas.ac.cn}

\author{Xiaoan Zhang}
\orcid{0009-0002-0659-851X}
\affiliation{%
  \institution{State Key Laboratory of Multimedia Information Processing, School of Computer Science, Peking University}
  \state{Beijing}
  \country{China}
}
\email{2401210538@stu.pku.edu.cn}

\author{Jianing Li}
\orcid{0000-0001-9552-2247}
\affiliation{%
  \institution{School of electronic science and engineering, Nanjing University}
  \city{Nanjing}
  \state{Jiangsu}
  \country{China}
}
\email{jnli@smail.nju.edu.cn}

\author{Chengyu Bai}
\orcid{0009-0007-6937-0877}
\affiliation{%
  \institution{State Key Laboratory of Multimedia Information Processing, School of Computer Science, Peking University}
  \state{Beijing}
  \country{China}
}
\email{2401210567@stu.pku.edu.cn}

\author{Ying Li}
\orcid{0009-0004-2587-3489}
\affiliation{%
  \institution{State Key Laboratory of Multimedia Information Processing, School of Computer Science, Peking University}
  \state{Beijing}
  \country{China}
}
\email{2301210309@stu.pku.edu.cn}

\author{Ming Lu}
\orcid{0000-0001-6819-6490}
\affiliation{%
  \institution{State Key Laboratory of Multimedia Information Processing, School of Computer Science, Peking University}
  \state{Beijing}
  \country{China}
}
\email{lu199192@gmail.com}

\author{Wenzhao Zheng}
\orcid{0000-0001-7188-3734}
\affiliation{%
  \institution{Berkeley Artificial Intelligence Research Lab, Department of EECS, University of California, Berkeley}
  \city{Berkeley}
  \state{California}
  \country{USA}
}
\email{wzzheng@berkeley.edu}

\author{Shanghang Zhang}
\authornote{Corresponding author.}
\orcid{0000-0003-4047-3526}
\affiliation{%
  \institution{State Key Laboratory of Multimedia Information Processing, School of Computer Science, Peking University}
  \state{Beijing}
  \country{China}
}
\email{shanghang@pku.edu.cn}

\renewcommand{\shortauthors}{Hao Wang et al.}

\begin{abstract}
Online 3D occupancy prediction provides a comprehensive spatial understanding of embodied environments. While the innovative EmbodiedOcc framework utilizes 3D semantic Gaussians for progressive indoor occupancy prediction, it overlooks the geometric characteristics of indoor environments, which are primarily characterized by planar structures. This paper introduces EmbodiedOcc++, enhancing the original framework with two key innovations: a Geometry-guided Refinement Module (GRM) that constrains Gaussian updates through plane regularization, along with a Semantic-aware Uncertainty Sampler (SUS) that enables more effective updates in overlapping regions between consecutive frames. GRM regularizes the position update to align with surface normals. It determines the adaptive regularization weight using curvature-based and depth-based constraints, allowing semantic Gaussians to align accurately with planar surfaces while adapting in complex regions. To effectively improve geometric consistency from different views, SUS adaptively selects proper Gaussians to update. Comprehensive experiments on the EmbodiedOcc-ScanNet benchmark demonstrate that EmbodiedOcc++ achieves state-of-the-art performance across different settings. Our method demonstrates improved edge accuracy and retains more geometric details while ensuring computational efficiency, which is essential for online embodied perception. The code will be released at: https://github.com/PKUHaoWang/EmbodiedOcc2.
\end{abstract}



\begin{teaserfigure}
  \includegraphics[width=\textwidth]{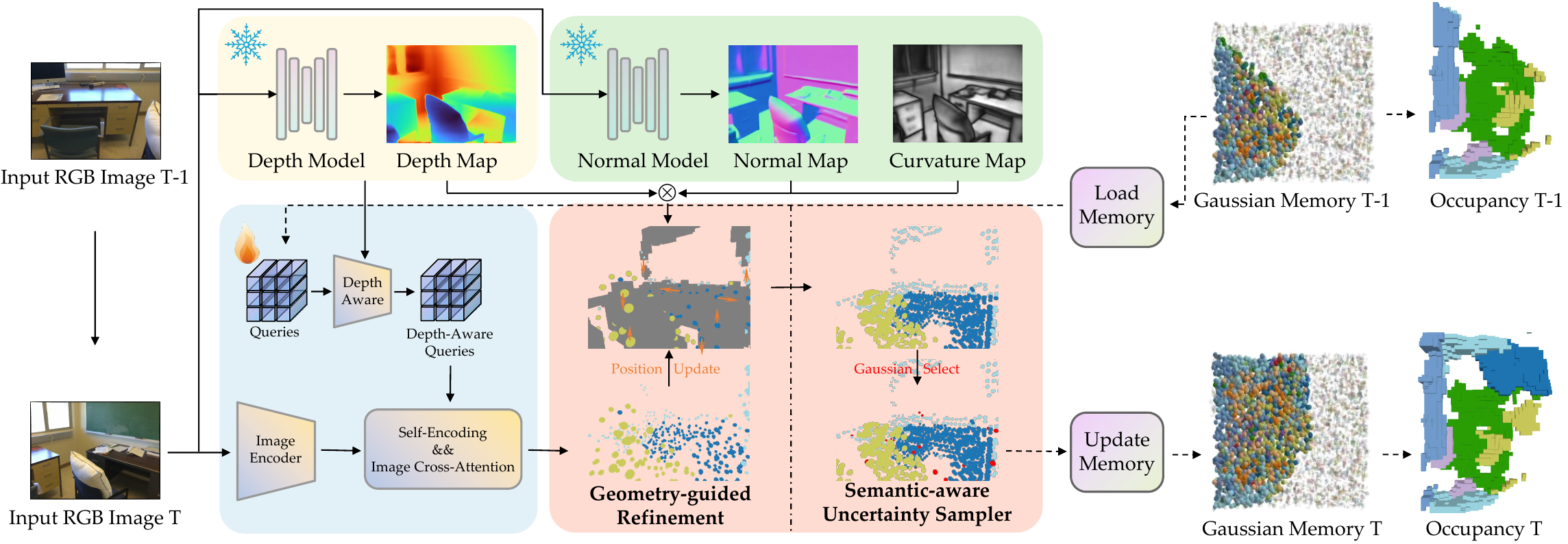}
  \caption{EmbodiedOcc++ Framework: A monocular RGB-based indoor 3D occupancy prediction method enhanced with plane regularization and uncertainty sampling. The Geometry-guided Refinement Module regularizes Gaussian updates along surface tangent planes using adaptive curvature and depth constraints, while the Semantic-aware Uncertainty Sampler enables efficient Gaussian selection. Designed for embodied scene understanding with preserved planar structures during progressive indoor exploration.
  }
  % \vspace{-2mm}
  \label{fig:teaser}
\end{teaserfigure}


\maketitle

\section{Introduction}
3D scene understanding has emerged as a fundamental challenge in computer vision, playing a crucial role across numerous applications, including robotics navigation, augmented reality, and autonomous driving~\cite{humblot2022navigation, zhang2022outdoor, li2022interactive, kong2025multi, li2023mseg3d, wei2024nto3d, huang2024textit, wei2024emd}. Among various 3D perception approaches, occupancy prediction~\cite{cao2022monoscene, huang2023tri, huang2024gaussianformer, wei2023surroundocc, yu2024monocular, wang2023openoccupancy, zhang2023occformer, li2025sliceocc} has gained significant traction due to its comprehensive representation capabilities. 
Unlike 3D object detection, which relies on bounding boxes and overlooks geometric details, occupancy prediction represents scenes as semantically labeled voxels. This approach captures detailed structures and enhances understanding for tasks such as planning. Its voxel-based representation effectively accommodates objects of different sizes and shapes, making it ideal for complex real-world environments~\cite{mescheder2019occupancy, niemeyer2019occupancy}. 

Most existing methods for predicting 3D occupancy are designed for outdoor autonomous driving and can be categorized into three main groups. Planar-based methods~\cite{huang2023tri, zuo2023pointocc, hou2024fastocc, li2025sliceocc}, such as those utilizing BEV (Bird's Eye View) or TPV (Tri-Perspective View), project features onto planes to reduce computational complexity. 
Voxel-based methods divide the 3D space into regular grid cells, performing 2D-to-3D lifting~\cite{zhang2023occformer, tang2024sparseocc} or point voxelization~\cite{wang2023openoccupancy} to extract features using 3D convolutions. 
Although thorough, these methods often face inefficiencies from processing empty voxels. 
More recently, Gaussian-based methods like GaussianFormer~\cite{huang2024gaussianformer} and its extensions~\cite{huang2024probabilistic, wu2024embodiedocc} represent scenes using 3D semantic Gaussians, addressing sparsity through an object-centric design in which each Gaussian encodes semantic features over a flexible region, demonstrating significant potential for 3D occupancy prediction.
    
However, adapting these approaches to indoor environments presents new challenges. 
EmbodiedOcc~\cite{wu2024embodiedocc} builds upon GaussianFormer~\cite{huang2024gaussianformer} to enable online indoor occupancy prediction using memory modules. However, it lacks designs specific to indoor environments and overlooks essential geometric characteristics of embodied spaces. 
The challenges for indoor occupancy prediction lie in two-fold:
(1) \textbf{Distinctive geometric characteristics.} Indoor environments show distinct geometric patterns, especially with planar surfaces like walls, floors, and furniture, unlike outdoor settings where objects follow structured pathways. Additionally, indoor scenes often contain many tightly packed objects in small spaces, leading to frequent occlusions and complex spatial relationships. However, current methods overlook geometric characteristics, resulting in blurred details and imprecise boundaries.
(2) \textbf{Redundancy in memory updates.} Embodied occupancy prediction requires online updates as agents continuously observe the indoor environment. The memory stores 3D semantic Gaussians as the scene representation, which are incrementally updated over time. However, overlapping observations from consecutive frames often result in redundant updates, which degrade the quality of Gaussian refinement.  
Existing methods simply treat all observations equally, leading to noisy Gaussian updates and redundant computation. 

\begin{figure}[!t]
  \includegraphics[width=0.49\textwidth]{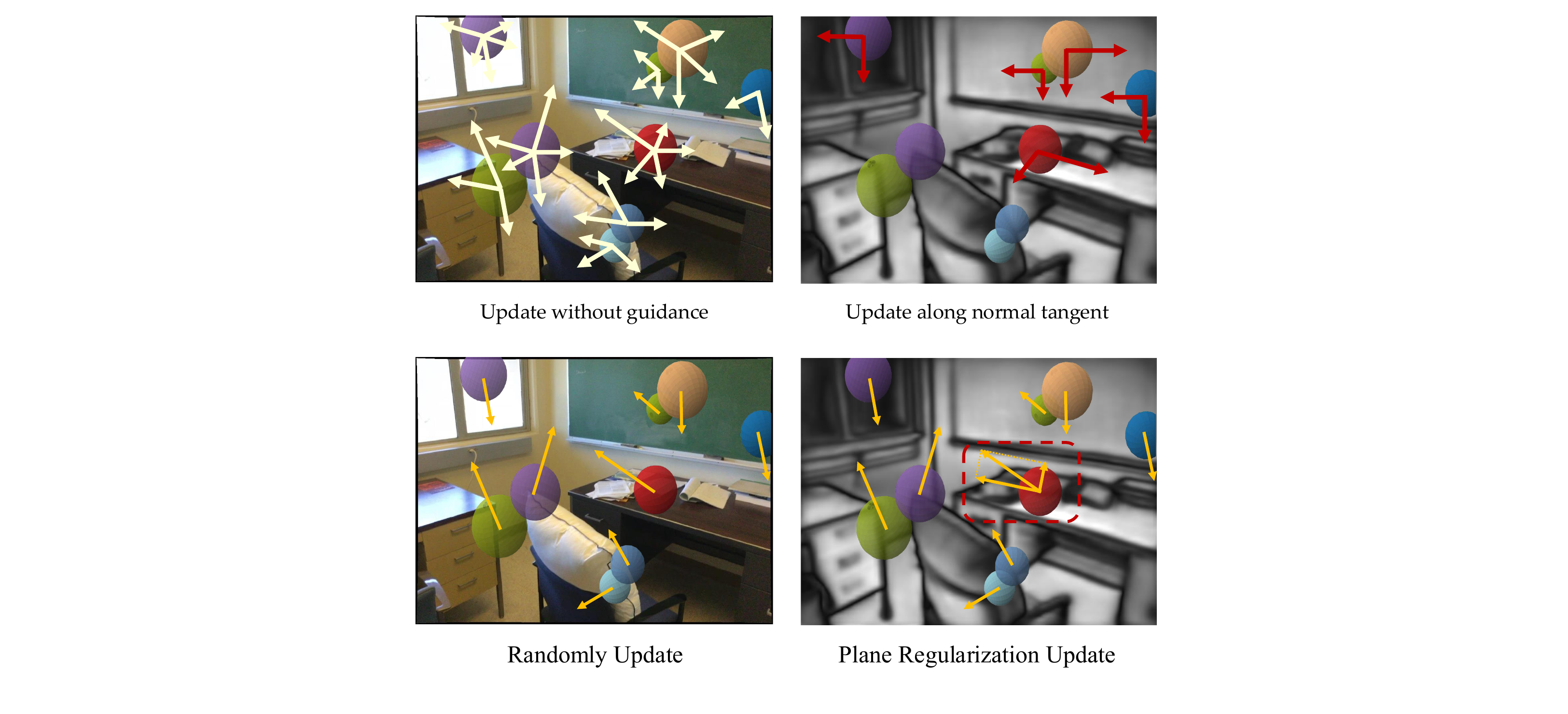}
  \vspace{-6mm}
  \caption{Illustration of different Gaussian updates. Left: The previous approach involved randomly updating positions without applying plane regularization. Right: Our approach employs plane regularization to refine Gaussian positions, leading to more precise representations of indoor scenes.}
  \label{fig:motivation}
  \vspace{-6mm}
\end{figure}

To address these limitations, we propose EmbodiedOcc++  (Fig.~\ref{fig:teaser}), an enhanced framework dedicated to improving embodied occupancy prediction by leveraging planar regularization and uncertainty sampler in indoor environments. 
First, to leverage the distinctive geometric characteristics of indoor scenes, we introduce a Geometry-guided Refinement Module that constrains the position updates of 3D semantic Gaussians (Fig.~\ref{fig:motivation}). 
By regularizing the position updates primarily along the tangent plane of surface normal, our approach explicitly regularizes Gaussian refinement to align with dominant planar structures in indoor environments. 
To preserve sharp edges and structural boundaries, we adaptively weigh the planar regularization. Specifically, stronger regularization is applied in flat regions dominated by planar structures, while looser constraints are used in areas with high curvature or complex geometry. 
Second, to mitigate redundancy in memory updates, we propose a Semantic-aware Uncertainty Sampler that adaptively selects low-confidence Gaussians for subsequent updates. 
Unlike EmbodiedOcc~\cite{wu2024embodiedocc}, which applies uniform weights when updating Gaussians in overlapping regions, our method estimates uncertainty for each Gaussian based on its semantic information. 
This uncertainty estimation enables differential weighting of Gaussians across consecutive frames, allowing robust and efficient memory updates in overlapping regions and improving geometric consistency for indoor occupancy prediction.

Without introducing additional trainable parameters, EmbodiedOcc++ maintains the progressive updating mechanism while significantly enhancing the evolution of Gaussian distributions. 
Our adaptive geometric constraint technique encourages Gaussian to better conform to planar surfaces, resulting in a more accurate representation of structural elements such as walls, floors, and furniture. Our main contributions are as follows:

\begin{itemize}
\item We propose a Geometry-guided Refinement Module (GRM) that constrains Gaussian updates through plane regularization, adaptively enforcing strong constraints only when both curvature and depth cues indicate planar regions.  
\item We propose a Semantic-aware Uncertainty Sampler (SUS) that adaptively selects and updates low-confidence Gaussians in overlapping regions between consecutive frames to mitigate redundancy in memory updates.
\item Our method achieves state-of-the-art (SOTA) performance on the EmbodiedOcc-ScanNet benchmark across various indoor occupancy prediction settings. 
\end{itemize}

\section{Related Work}
\subsection{Indoor Neural Representation}
Neural representation for indoor 3D scenes has evolved along several research directions. Early approaches focused on extracting 3D meshes using voxel volumes~\cite{peng2020convolutional, sun2021neuralrecon, stier2021vortx} and TSDF-fusion~\cite{sayed2022simplerecon}, which offer efficient mesh extraction but sacrifice photorealistic neural rendering. SLAM-based methods~\cite{zhu2022nice, yan2024gs, keetha2024splatam} use dense RGB-D input for real-time mapping but struggle to scale in dynamic scenes. 
Recent methods have explored implicit representations like Signed Distance Fields~\cite{yariv2021volume, yu2022monosdf, li2023neuralangelo, wei2024medsam}, which learn powerful scene representations but require intensive per-scene optimization. SurfelNeRF~\cite{gao2023surfelnerf} maps image sequences to 3D surfels in a feed-forward manner, yet suffers from slow optimization.
Our work EmbodiedOcc++ differs by utilizing an efficient 3D Gaussian representation with geometry-guided refinement specifically designed for indoor environments. 

\subsection{Occupancy Prediction}
Occupancy prediction enhances multimodal perception capabilities by dividing scenes into semantic 3D grids. MonoScene~\cite{cao2022monoscene} pioneers deriving occupancy from single images, while subsequent works~\cite{yao2023ndc, yu2024monocular} address depth ambiguity challenges. Current methods fall into three categories: Planar-based approaches like TPVFormer~\cite{huang2023tri} and SliceOcc~\cite{li2025sliceocc} project features onto orthogonal planes. Voxel-based methods such as SurroundOcc~\cite{wei2023surroundocc} and OpenOccupancy~\cite{wang2023openoccupancy} obtain voxel features through image-volume cross-attention or by lifting image features into 3D space. Several approaches~\cite{pan2024renderocc, huang2024selfocc, zhao2024hybridocc} combine multimodal priors for novel view rendering. 
GaussianFormer~\cite{huang2024gaussianformer, huang2024probabilistic} introduces an object-centric approach using sparse 3D semantic Gaussians, achieving comparable performance with reduced memory usage. 
Following progress in outdoor environments, recent work turns to occupancy prediction for indoor scenes. 
EmbodiedScan~\cite{wang2024embodiedscan} establishes a benchmark for indoor occupancy prediction, while ISO~\cite{yu2024monocular} tackles indoor challenges using monocular images and depth maps. 
EmbodiedOcc~\cite{wu2024embodiedocc} further enables online occupancy prediction. 
However, these methods overlook the intricate geometry of embodied scenes. In contrast, our EmbodiedOcc++ is the first to account for the planar structures that are prevalent in indoor environments. 

\vspace{-2mm}
\subsection{Indoor 3D Gaussian Splatting}
3D Gaussian Splatting (3DGS)~\cite{kerbl20233d} and its variants~\cite{szymanowicz2024splatter, qian2024gaussianavatars, lin2024vastgaussian, wei2024gazegaussian, wei2024graphavatar, wang2024plgs} have shown promising rendering results and faster performance compared to NeRF~\cite{mildenhall2021nerf}. 
However, vanilla optimization for 3DGS leads to disorganized Gaussian distributions due to the absence of explicit geometric constraints, resulting in poor surface continuity and imprecise object boundaries. 
Several approaches address these limitations, including DN-Splatter~\cite{turkulainen2024dn}, GSDF~\cite{yu2024gsdf} and OmniIndoor3D~\cite{wei2025omniindoor3d}, which incorporate geometric priors, while SuGaR~\cite{guedon2024sugar}, PGSR~\cite{chen2024pgsr}, and RaDe-GS~\cite{zhang2024rade} employ flattened Gaussians to improve surface reconstruction. 

However, these methods focus on per-scene optimization and fail to incorporate geometric constraints into occupancy prediction. 
To the best of our knowledge, we are the first to investigate geometry-guided Gaussian refinement in embodied environments for online occupancy prediction. We further propose a novel semantic-aware uncertainty sampler that adaptively refines Gaussians.

\section{Methodology}
\subsection{Embodied Occupancy Prediction} 
As the foundation of our framework, we utilize EmbodiedOcc~\cite{wu2024embodiedocc}, which conducts online 3D occupancy prediction in indoor environments using semantic Gaussians. It allows for progressive refinement of scenes as the embodied agent navigates the environment. Next, we provide a brief overview of the core modules in EmbodiedOcc. 

\textbf{Local Occupancy Prediction.} The local occupancy prediction module uses monocular RGB input to predict 3D occupancy within the current camera frustum through a sophisticated Gaussian-based representation. Each Gaussian is characterized by its position $\mathbf{m} \in \mathbb{R}^3 $, scale $\mathbf{s} \in \mathbb{R}^3$, rotation quaternion $\mathbf{r} \in \mathbb{R}^4 $, opacity $\mathbf{o} \in \mathbb{R}^1 $, and semantic logits $\mathbf{c} \in \mathbb{R}^{12} $, operating within the camera coordinate system. A set of semantic Gaussians is initialized in the frustum and refined through a multi-stage pipeline: first, feature vectors are updated via a self-encoder and image cross-attention; then, the Gaussians are refined using the following equation: 
\begin{equation}
    \mathbf{G}_{new} = (\Delta \mathbf{m}+\mathbf{m}, \Delta \mathbf{s}+\mathbf{s}, \Delta \mathbf{r} \otimes \mathbf{r}, \Delta \mathbf{o}+\mathbf{o}, \Delta \mathbf{c}+\mathbf{c}),
\end{equation}
where each Gaussian attribute is updated by adding its corresponding residual. Then, Gaussian-to-Voxel Splatting transforms refined Gaussians into the predictive occupancy within the camera frustum.

\textbf{Gaussian Memory.} The Gaussian memory acts as a global scene representation in world coordinates, initially distributed uniformly and updated continuously as the agent explores. At each time step $t$, the framework receives a posed visual input $\mathit{x}_{t} = (\mathit{I}_{t}, \mathit{M}_{t})$ and updates the Gaussians within the current frustum.
Each Gaussian is updated using a fixed weight determined at initialization, with higher-weighted Gaussians undergoing finer refinements and lower-weighted ones receiving more aggressive updates. This strategy progressively improves accuracy as exploration proceeds.

\textbf{Embodied Framework.} The EmbodiedOcc framework combines the local occupancy prediction module with Gaussian memory to enable embodied 3D occupancy prediction. 
The training pipeline proceeds through two primary stages: first, the local occupancy prediction module is trained using monocular inputs and ground truth data, and subsequently, the Gaussian memory is initialized and iteratively updated with losses computed after each update to maintain scene-wide coherence. The model is optimized with a combination of focal loss, Lovasz-softmax loss, and scene-class affinity losses.
During exploration, the framework updates scene representations incrementally while maintaining consistency in previously explored areas through a memory-based mechanism.

\subsection{Geometry-guided Refinement Module }
To address random Gaussian updates in EmbodiedOcc~\cite{wu2024embodiedocc}, 
we introduce a novel Geometry-guided Refinement Module that leverages geometric cues from monocular input to better model the predominantly planar structures in indoor environments. This module incorporates constraints that guide Gaussians to align with planar surfaces, enhancing geometric details.

\begin{figure*}
  \includegraphics[width=\textwidth]{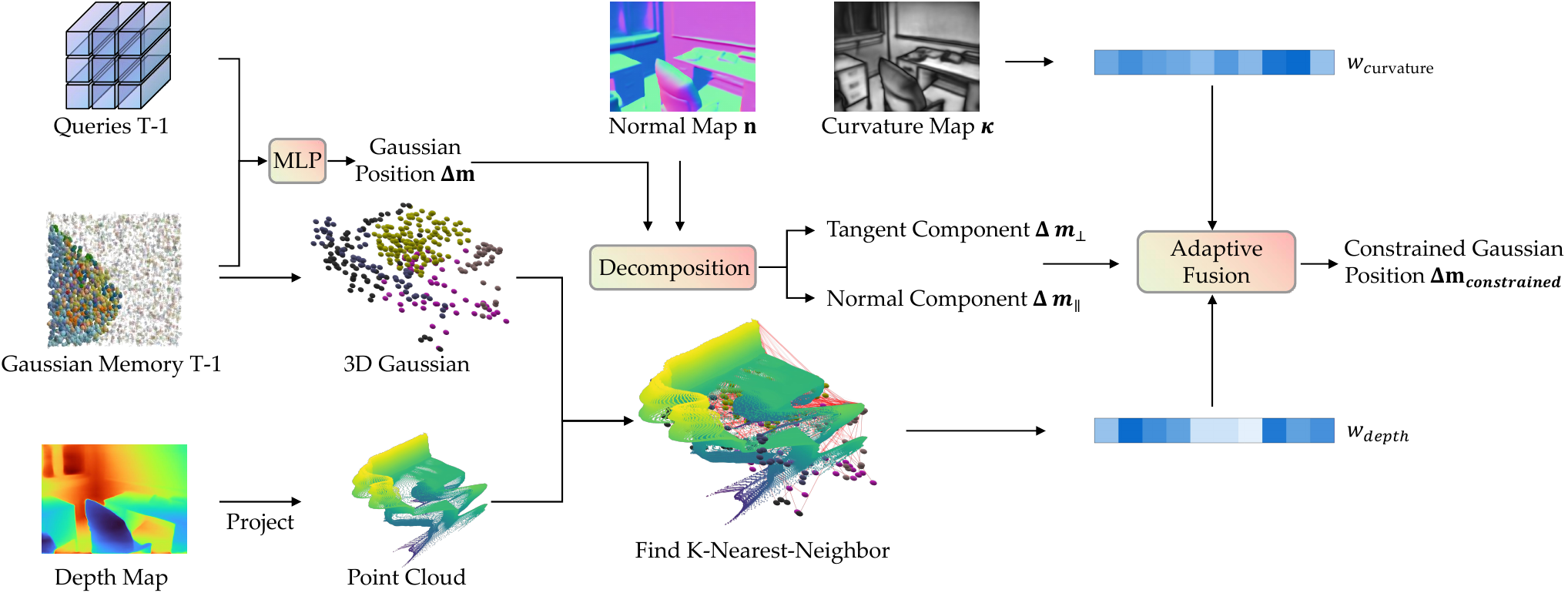}
  \vspace{-6mm}
  \caption{Geometry-guided Refinement Module. Our approach leverages monocular geometric cues to constrain Gaussian updates, decomposing position updates into components parallel and perpendicular to estimated surface normals. The module incorporates both surface curvature and depth information to dynamically adjust constraint weights, allowing stronger planar constraints in flat regions while preserving flexibility in areas with complex geometry.}
  % \Description{2.}
  \vspace{-4mm}
  \label{fig:refine}
\end{figure*}

\textbf{Monocular Cues constrained Optimization.}
The core innovation of our approach lies in the geometric refinement of Gaussians during their position updates. While the original EmbodiedOcc framework updates Gaussian parameters through a general delta refinement process, our method introduces specialized constraints guided by geometric cues derived from monocular input. We leverage a pre-trained model~\cite{bae2021estimating} to obtain monocular normal priors from the input images. During the refinement process, we only consider those Gaussians that are visible within the current camera view, ensuring that our geometric constraints are applied only to points with reliable visual information. 

When refining the position of Gaussians, we decompose the position update vector $\Delta \mathbf{m}$ into components parallel $\Delta\mathbf{m}_{\parallel}$ and perpendicular $\Delta\mathbf{m}_{\perp}$ to the estimated surface normal:
\begin{equation}
    \begin{aligned}
        \Delta\mathbf{m}_{\parallel} &= (\Delta\mathbf{m} \cdot \mathbf{n}) \mathbf{n} \\
        \Delta\mathbf{m}_{\perp} &= \Delta\mathbf{m} - \Delta\mathbf{m}_{\parallel},
    \end{aligned}
\end{equation}
where $\mathbf{n}$ represents the surface normal at the Gaussian's projected location in the image. The constrained position update is then calculated as:
\begin{equation}
    \Delta\mathbf{m}_{constrained}=w\cdot\Delta\mathbf{m}_\perp+(1-w)\cdot\Delta\mathbf{m},
\end{equation}
where $w$ is the normal constraint weight that determines the strength of the planar constraint. This formulation encourages Gaussians to move primarily along the tangent plane, preserving the planar structure of indoor environments while still allowing for necessary adjustments in all directions.

\textbf{Surface Curvature based Constraint.} 
To adapt the planar constraints according to local surface properties, we introduce a curvature-guided weighting mechanism. The curvature map, which represents local surface curvature, is obtained using the same pre-trained model~\cite{bae2021estimating} that provides our normal priors. This curvature map provides crucial information about where strong planar constraints should be applied (low curvature regions) versus where more flexible updates are needed (high curvature regions). 

For each valid Gaussian, we determine its corresponding curvature by projecting it onto the image plane and extracting the curvature value at that location. The normal constraint weight is then dynamically adjusted based on this curvature value: 
\begin{equation}
    w_{\kappa} =
    \begin{cases}
        w_{min} & \text{if } \kappa \leq \kappa_{low} \\
        w_{min} + \frac{\kappa - \kappa_{low}}{\kappa_{high} - \kappa_{low}} \cdot (w_{max} - w_{min}) & \text{if } \kappa_{low} < \kappa < \kappa_{high} \\
        w_{max} & \text{if } \kappa \geq \kappa_{high},
    \end{cases}
\end{equation}
where $\kappa_{low}$ and $\kappa_{high}$ are threshold values that define the transition between low and high curvature regions, and $w_{min}$ and $w_{max}$ are the minimum and maximum normal constraint weights. This approach allows for stronger planar constraints in flat regions while preserving flexibility in areas with complex geometry.

\textbf{Depth-aware Spatial Constraint.}
In addition to curvature information, we leverage depth cues to refine the planar constraints. We utilize the fine-tuned DepthAnything-V2 model~\cite{yang2024depth} to obtain high-quality depth maps from our monocular input. By converting the predicted depth map into a point cloud, we establish a spatial relationship between Gaussians and nearby point clouds.

For each valid Gaussian, we compute its distance to the nearest depth point in 3D space. This distance serves as an indicator of confidence in the point's position relative to the observed surface:
\begin{equation}
    w_{depth}=\mathrm{clamp}\left(\frac{d_{far}-d_{min}}{d_{far}-d_{near}},0,1\right),
\end{equation}
where $d_{min}$ is the distance to the nearest depth point, and $d_{near}$ and $d_{far}$ are threshold values. This weighting scheme applies stronger planar constraints to Gaussians that are close to observed surfaces and relaxes constraints for points in regions with sparse or uncertain depth information.

\textbf{Adaptive Constraint Fusion.}
To leverage the complementary strengths of both curvature and depth-based constraints, we introduce an adaptive fusion mechanism that combines these cues into a single, more robust constraint weight. In our implementation, we specifically adopt a product fusion strategy:
\begin{equation}
    w_{fused}=w_{depth} \cdot w_{curvature},
\end{equation}
This multiplication-based fusion ensures that the final constraint is strong only when both curvature and depth cues agree on the presence of a planar structure. The product operation acts as a logical "AND" between the two constraints—a point must satisfy both the depth proximity criterion and the low curvature requirement to receive a strong planar constraint. This conservative approach prevents over-constraining in uncertain regions and naturally handles scenarios where one cue might be unreliable. For instance, in texture-less areas where depth estimation might be challenging or in visually complex regions where curvature estimation might be noisy, the combined weight will appropriately reduce the strength of the planar constraint.

By integrating these geometric constraints into the Gaussian refinement process through product fusion, our approach significantly improves the representation of planar structures common in indoor environments, resulting in more accurate and visually coherent scene reconstructions.

\subsection{Semantic-aware Uncertainty Sampler}
To address the redundancy in memory updates caused by overlapping observations from consecutive frames, we propose a Semantic-aware Uncertainty Sampler. In contrast to fixed update weights in EmbodiedOcc~\cite{wu2024embodiedocc}, this module estimates the uncertainty of each Gaussian semantic representation and adaptively selects low-confidence Gaussians for further refinement.

\textbf{Monte Carlo Sampling for Uncertainty Estimation.}
Our approach employs Monte Carlo dropout sampling~\cite{gal2016dropout} to capture the epistemic uncertainty in semantic predictions. Rather than relying on a single deterministic prediction, we perform multiple forward passes with dropout enabled during inference:
\begin{equation}
    \Delta\mathbf{c}_i = f_\theta(\mathbf{x} + \mathbf{e}_i), \quad i = 1,2,...,M,
\end{equation}
where $\Delta\mathbf{c}_i$ represents the semantic delta prediction for the $i$-th sample, $f_\theta$ is our feature extraction network with parameters $\theta$, and $\mathbf{e}_i$ indicates the effective noise introduced by dropout. We use $M=3$ samples in our implementation to balance computational efficiency and reliable uncertainty estimation.

\textbf{Entropy-based Uncertainty Quantification.}
After obtaining multiple semantic predictions through Monte Carlo sampling, we compute the mean semantic distribution by combining the predicted deltas with the original semantics:
\begin{equation}
    \mathbf{p} = \frac{1}{M}\sum_{i=1}^{M}\text{softmax}(\Delta\mathbf{c}_i + \mathbf{c}),
\end{equation}
where $\mathbf{c}$ represents the original semantic features.

The uncertainty of each Gaussian is then quantified using the entropy of this mean distribution:
\begin{equation}
    H(\mathbf{p}) = -\sum_{j=1}^{C} p_j \log p_j,
\end{equation}
where $C$ is the number of semantic classes and $p_j$ is the probability of class $j$. To ensure comparability across different numbers of semantic classes, we normalize the entropy:
\begin{equation}
    \hat{H}(\mathbf{p}) = \frac{H(\mathbf{p})}{\log C},
\end{equation}
This normalized entropy ranges from 0 to 1, where higher values indicate greater uncertainty in the semantic prediction.

\textbf{Uncertainty-Guided Gaussian Update.}
The estimated uncertainty is crucial in determining how Gaussians should be updated when the same region is observed from different viewpoints. We directly utilize the normalized entropy as the update ratio:
\begin{equation}
    r = \hat{H}(\mathbf{p}),
\end{equation}
which controls how much new information should be incorporated from new observations. For regions with high semantic uncertainty, we allow greater updates from new observations, while points with low uncertainty maintain more of their original properties. 

\definecolor{ceiling}{RGB}{214,  38, 40}   %
\definecolor{floor}{RGB}{43, 160, 4}     %
\definecolor{wall}{RGB}{158, 216, 229}  %
\definecolor{window}{RGB}{114, 158, 206}  %
\definecolor{chair}{RGB}{204, 204, 91}   %
\definecolor{bed}{RGB}{255, 186, 119}  %
\definecolor{sofa}{RGB}{147, 102, 188}  %
\definecolor{table}{RGB}{30, 119, 181}   %
\definecolor{tvs}{RGB}{160, 188, 33}   %
\definecolor{furniture}{RGB}{255, 127, 12}  %
\definecolor{objects}{RGB}{196, 175, 214} %

\begin{table*}[!ht]
		\caption{
        \textbf{Local prediction performance on the Occ-ScanNet datasets. }
        }
        \vspace{-5mm}
        % \vspace{-7mm}
		\small
		\setlength{\tabcolsep}{0.008\textwidth}
		\captionsetup{font=scriptsize}
            \begin{center}
            \resizebox{1.0\linewidth}{!}{
		\begin{tabular}{l|c|c|c c c c c c c c c c c|c}
			\toprule
			Dataset
			& Method
			& {IoU}
			& \rotatebox{90}{\parbox{1.5cm}{\textcolor{ceiling}{$\blacksquare$} ceiling}} 
			& \rotatebox{90}{\textcolor{floor}{$\blacksquare$} floor}
			& \rotatebox{90}{\textcolor{wall}{$\blacksquare$} wall} 
			& \rotatebox{90}{\textcolor{window}{$\blacksquare$} window} 
			& \rotatebox{90}{\textcolor{chair}{$\blacksquare$} chair} 
			& \rotatebox{90}{\textcolor{bed}{$\blacksquare$} bed} 
			& \rotatebox{90}{\textcolor{sofa}{$\blacksquare$} sofa} 
			& \rotatebox{90}{\textcolor{table}{$\blacksquare$} table} 
			& \rotatebox{90}{\textcolor{tvs}{$\blacksquare$} tvs} 
			& \rotatebox{90}{\textcolor{furniture}{$\blacksquare$} furniture} 
			& \rotatebox{90}{\textcolor{objects}{$\blacksquare$} objects} 
			& mIoU\\
			\midrule
			\multirow{4}{*}{Occ-ScanNet-mini} 
			& MonoScene~\cite{cao2022monoscene} & 0.419 & 0.170 & 0.462 & 0.239 & 0.127 & 0.270 & 0.291 & 0.348 & 0.291 & 0.097 & 0.345 & 0.204 & 0.259 \\
            & ISO~\cite{yu2024monocular} & 0.429 & 0.211 & 0.427 & 0.246 & 0.151 & 0.308 & 0.410 & 0.433 & 0.322 & 0.121 & 0.359 & 0.251 & 0.294 \\
            & EmbodiedOcc~\cite{wu2024embodiedocc} & 0.538 & \textbf{0.291} & 0.487 & 0.423 & 0.387 & 0.420 & 0.627 & 0.606 & 0.482 & 0.338 & 0.580 & 0.465 & 0.464 \\
            & EmbodiedOcc++ & \textbf{0.557} & 0.233 & \textbf{0.510} & \textbf{0.428} & \textbf{0.393} & \textbf{0.435} & \textbf{0.656} & \textbf{0.640} & \textbf{0.507} & \textbf{0.407} & \textbf{0.603} & \textbf{0.489} & \textbf{0.482} \\
            \midrule
			\multirow{4}{*}{Occ-ScanNet}
			& MonoScene~\cite{cao2022monoscene} & 0.416 & 0.152 & 0.447 & 0.224 & 0.126 & 0.261 & 0.270 & 0.359 & 0.283 & 0.066 & 0.322 & 0.198 & 0.246 \\
            & ISO~\cite{yu2024monocular} & 0.422 & 0.199 & 0.419 & 0.224 & 0.170 & 0.291 & 0.424 & 0.420 & 0.296 & 0.106 & 0.364 & 0.246 & 0.287 \\
            & EmbodiedOcc~\cite{wu2024embodiedocc} & 0.539 & \textbf{0.409} & 0.508 & \textbf{0.419} & 0.330 & 0.412 & 0.552 & 0.619 & 0.438 & \textbf{0.354} & 0.535 & \textbf{0.429} & 0.455 \\
            & EmbodiedOcc++ & \textbf{0.549} & 0.364 & \textbf{0.531} & 0.418 & \textbf{0.344} & \textbf{0.429} & \textbf{0.573} & \textbf{0.641} & \textbf{0.452} & 0.348 & \textbf{0.542} & 0.441 & \textbf{0.462} \\
			\bottomrule
		\end{tabular}}
            \end{center}
% \vspace{-7mm}
		\label{tab:mono}
        \vspace{-5mm}
 \end{table*}

To further improve computational efficiency, we introduce an uncertainty threshold $\tau_{unc}$. Gaussians with uncertainty below this threshold are considered sufficiently reliable and are excluded from subsequent updates:
\begin{equation}
    r =
    \begin{cases}
        \hat{H}(\mathbf{p}) & \text{if } \hat{H}(\mathbf{p}) > \tau_{unc} \\
        0 & \text{otherwise},
    \end{cases}
\end{equation}
this thresholding mechanism significantly reduces the number of Gaussians that need to be processed in later iterations, leading to more efficient optimization compared to the original EmbodiedOcc framework~\cite{wu2024embodiedocc}, which updates all points in a fixed manner.

For points that do require updates, we apply the update ratio to all Gaussian properties during the update process:
\begin{equation}
    \mathbf{G}_{updated} = r \cdot \Delta\mathbf{G} + (1-r) \cdot \mathbf{G},
\end{equation}
where $\mathbf{G}$ represents the original Gaussian properties, including mean position, scale, rotation, opacity, and semantic features, and $\Delta\mathbf{G}$ represents the predicted adjustments to these properties.

By incorporating the Semantic-aware Uncertainty Sampler, our approach enables adaptive and robust updates in overlapping regions between consecutive frames. This mechanism improves consistency when the same spatial region is observed from different viewpoints, mitigating the redundancy in memory updates.

%%%%%%%%%%%%%%%%%%%%%%%%%%%%%%%%%%%%%%%
\section{Experiments}
\subsection{Settings}
\noindent\textbf{Datasets. }Our model is evaluated on the EmbodiedOcc-ScanNet dataset~\cite{wu2024embodiedocc}, which is based on the Occ-ScanNet dataset~\cite{yu2024monocular}. This dataset provides monocular images paired with voxel-level semantic occupancy annotations. The dataset is divided into 537 training scenes and 137 validation scenes, with each scene comprising 30 posed frames and their corresponding occupancy annotations. In addition to the full dataset, EmbodiedOcc-ScanNet also provides mini versions: EmbodiedOcc-ScanNet-mini with 64/16 scenes in train/val splits respectively. The occupancy representation is structured as a $60 \times 60 \times 36$ voxel grid (corresponding to a $4.8m \times 4.8m \times 2.88m$ space in front of the camera) with a voxel resolution of $0.08m$. 
For global occupancy representations, the resolution varies based on scene dimensions and is calculated as $(lx \times ly \times lz)/0.08m$, where $lx, ly, lz$ represents the scene's spatial extent in world coordinates. The semantic annotations in this dataset encompass 12 distinct classes, comprising $11$ meaningful object and structural categories including architectural elements (ceiling, floor, wall, window), furniture (chair, bed, sofa, table), electronic devices (TVs), and general object classifications alongside an additional empty space class.

\noindent\textbf{Tasks. }Following EmbodiedOcc~\cite{wu2024embodiedocc}, we leverage the EmbodiedOcc-ScanNet dataset to evaluate our approach on two distinct tasks: local occupancy prediction and embodied occupancy prediction. For local occupancy prediction, we follow the established paradigm of using single monocular images to predict occupancy within the camera's frustum. For the challenging embodied occupancy prediction task, our method continuously processes sequential visual inputs to update occupancy estimates online. 

\noindent\textbf{Metrics. }Evaluation metrics encompass two performance indicators: the Scene Completion Intersection over Union (IoU) and the mean Intersection over Union (mIoU) for semantic scene understanding. The Scene Completion IoU provides a comprehensive metric for assessing the overall occupancy prediction accuracy, while the per-class mIoU offers detailed insights into the model's performance across different semantic categories. 
For local occupancy prediction, we strictly adhere to the ISO evaluation protocol~\cite{yu2024monocular}, computing these metrics within the camera frustum box. For embodied occupancy prediction, we expand our analysis to the global occupancy of each scene, focusing on regions comprehensively observed across the entire 30-frame sequence.

\begin{figure}
  \includegraphics[width=1.0\linewidth]{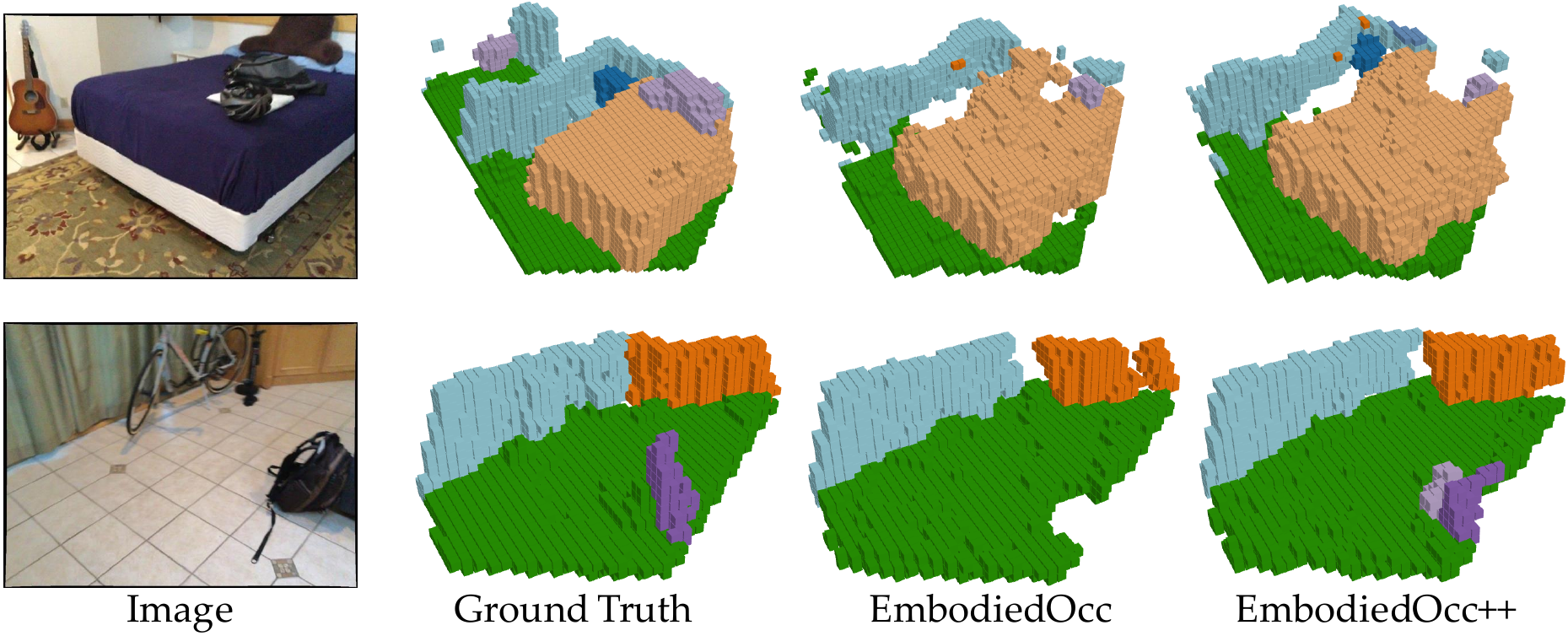}
  \vspace{-7mm}
  \caption{Visualization of our local occupancy prediction. Our method accurately reconstructs complex geometric structures and preserves fine details in challenging indoor scenes.}
  % \Description{2.}
  \label{fig:vis_mono}
  \vspace{-6mm}
\end{figure}

\definecolor{ceiling}{RGB}{214,  38, 40}   %
\definecolor{floor}{RGB}{43, 160, 4}     %
\definecolor{wall}{RGB}{158, 216, 229}  %
\definecolor{window}{RGB}{114, 158, 206}  %
\definecolor{chair}{RGB}{204, 204, 91}   %
\definecolor{bed}{RGB}{255, 186, 119}  %
\definecolor{sofa}{RGB}{147, 102, 188}  %
\definecolor{table}{RGB}{30, 119, 181}   %
\definecolor{tvs}{RGB}{160, 188, 33}   %
\definecolor{furniture}{RGB}{255, 127, 12}  %
\definecolor{objects}{RGB}{196, 175, 214} %
\begin{table*}[!ht]
		\caption{Embodied prediction performance on the EmbodiedOcc-ScanNet dataset.}
        \vspace{-4mm}
		\small
		\setlength{\tabcolsep}{0.008\textwidth}
		\captionsetup{font=scriptsize}
            \begin{center}
            \resizebox{1.0\linewidth}{!}{
		\begin{tabular}{l|l|c|c c c c c c c c c c c|c}
			\toprule
			Dataset
			& Method
			& {IoU}
			& \rotatebox{90}{\parbox{1.5cm}{\textcolor{ceiling}{$\blacksquare$} ceiling}} 
			& \rotatebox{90}{\textcolor{floor}{$\blacksquare$} floor}
			& \rotatebox{90}{\textcolor{wall}{$\blacksquare$} wall} 
			& \rotatebox{90}{\textcolor{window}{$\blacksquare$} window} 
			& \rotatebox{90}{\textcolor{chair}{$\blacksquare$} chair} 
			& \rotatebox{90}{\textcolor{bed}{$\blacksquare$} bed} 
			& \rotatebox{90}{\textcolor{sofa}{$\blacksquare$} sofa} 
			& \rotatebox{90}{\textcolor{table}{$\blacksquare$} table} 
			& \rotatebox{90}{\textcolor{tvs}{$\blacksquare$} tvs} 
			& \rotatebox{90}{\textcolor{furniture}{$\blacksquare$} furniture} 
			& \rotatebox{90}{\textcolor{objects}{$\blacksquare$} objects} 
			& mIoU\\
			\midrule
            \multirow{2}{*}{EmbodiedOcc-mini} 
            & SplicingOcc & 0.488 & \textbf{0.290} & 0.376 & 0.373 & 0.268 & 0.445 & \textbf{0.660} & 0.527 & 0.408 & 0.366 & 0.545 & 0.279 & 0.412 \\
            & EmbodiedOcc~\cite{wu2024embodiedocc} & 0.507 & 0.215 & \textbf{0.445} & 0.383 & 0.279 & 0.469 & 0.647 & \textbf{0.553} & 0.427 & 0.358 & 0.525 & 0.275 & 0.416 \\
            & EmbodiedOcc++ & \textbf{0.529} & 0.225 & 0.439 & \textbf{0.395} & \textbf{0.334} & \textbf{0.470} & 0.651 & 0.544 & \textbf{0.449} & \textbf{0.381} & \textbf{0.579} & \textbf{0.341} & \textbf{0.437} \\
            \midrule
		\multirow{2}{*}{EmbodiedOcc} 
		& SplicingOcc & 0.490 & \textbf{0.316} & 0.388 & 0.355 & 0.363 & 0.471 & 0.545 & 0.572 & 0.344 & 0.325 & 0.512 & 0.291 & 0.407 \\
            & EmbodiedOcc~\cite{wu2024embodiedocc} & 0.515 & 0.227 & \textbf{0.446} & 0.374 & 0.380 & \textbf{0.501} & 0.567 & \textbf{0.597} & 0.354 & \textbf{0.384} & 0.520 & 0.329 & 0.425 \\
            & EmbodiedOcc++ & \textbf{0.522} & 0.279 & 0.439 & \textbf{0.387} & \textbf{0.406} & 0.490 & \textbf{0.579} & 0.592 & \textbf{0.368} & 0.378 & \textbf{0.535} & \textbf{0.341} & \textbf{0.436} \\
			\bottomrule
		\end{tabular}}
            \end{center}
		\label{tab:online}
 \end{table*}

\begin{figure*}[!ht]
  \vspace{-4mm}
  \includegraphics[width=\textwidth]{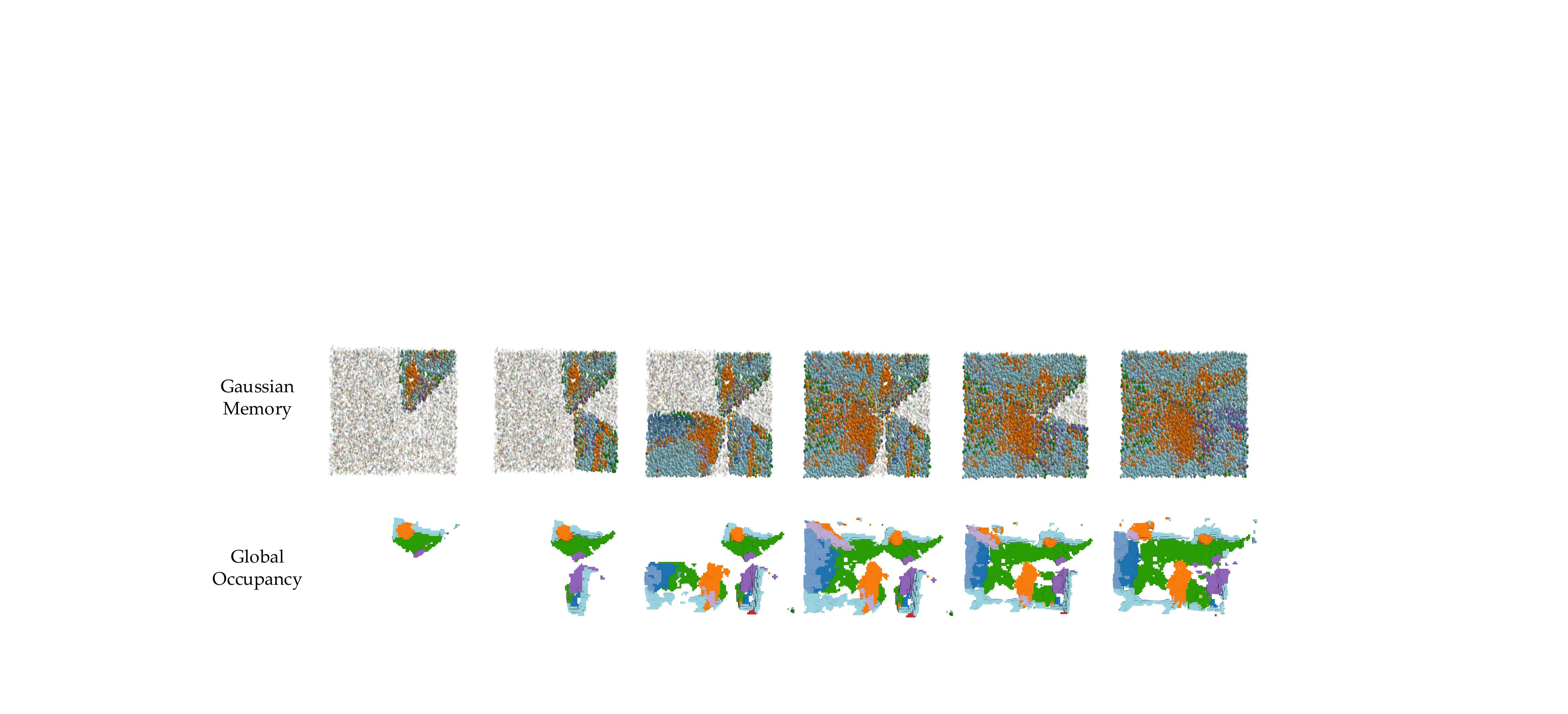}
  \vspace{-6mm}
  \caption{Visualization of our embodied occupancy prediction. Our method demonstrates superior performance in reconstructing complete 3D scenes by integrating information across multiple frames.}
  % \Description{2.}
  \label{fig:vis_embodied}
  \vspace{-4mm}
\end{figure*}

\vspace{-2mm}
\subsection{Implementation}
% \vspace{-1mm}
\noindent\textbf{Network Architecture. }
Our approach introduces a pre-trained model~\cite{bae2021estimating} to estimate normals and curvature values for each input image, which remains frozen during the training process. Additionally, we incorporate dropout layers into the Gaussian refinement module to effectively quantify uncertainty in our predictions. 
Without introducing additional trainable parameters, 
Without introducing additional trainable parameters, we follow the architectural design of EmbodiedOcc~\cite{wu2024embodiedocc} for all other components.

\noindent\textbf{Parameter Configuration. }
For our Geometry-guided Refinement module, we set curvature thresholds $\kappa_{low} = 5.0$ and $\kappa_{high} = 20.0$, normal constraint weights bounded by $w_{min} = 0.0$ and $w_{max} = 1.0$, distance thresholds $d_{near} = 0.1$ and $d_{far} = 0.25$, and employ $10$ neighbor points during nearest neighbor search. For our Semantic-aware Uncertainty Sampler used in embodied occupancy prediction, we initialize Gaussians with a $0.16m$ interval for scene representation. During each update, we set Monte Carlo dropout sampling times to $3$ and the uncertainty threshold to $0.3$ in the online refinement layer. We use the same loss functions and weighting scheme as EmbodiedOcc~\cite{wu2024embodiedocc}. 

\noindent\textbf{Optimization Strategy. }
We use an AdamW optimizer with 0.01 weight decay and a warm-up strategy for the first 1,000 iterations, followed by a cosine schedule. For local occupancy prediction, we train for 20 epochs on Occ-ScanNet-mini and 10 epochs on Occ-ScanNet using 8 NVIDIA A800 GPUs with a maximum learning rate of 2e-4. For embodied occupancy prediction, we train for 20 epochs on EmbodiedOcc-ScanNet-mini using 4 A800 GPUs with learning rate 1e-4, and 5 epochs on EmbodiedOcc-ScanNet using 8 GPUs with learning rate 2e-4.

\vspace{-7mm}
\subsection{Main Results}
% \vspace{-0.5mm}
\textbf{Local Occupancy Prediction.} 
We evaluate our proposed EmbodiedOcc++ against existing approaches for local occupancy prediction on both Occ-ScanNet-mini and Occ-ScanNet datasets. As shown in Tab.~\ref{tab:mono}, EmbodiedOcc++ consistently outperforms previous methods across multiple metrics. 
The consistent improvement across both mini and base dataset versions demonstrates that our approach scales effectively to larger and more diverse scene collections while maintaining its performance advantages. On object-level categories such as \textit{bed}, \textit{sofa}, \textit{table}, and \textit{furniture}, our method shows notable improvements, demonstrating its ability to better capture complex geometric structures and preserve object boundaries. Furthermore, EmbodiedOcc++ achieves superior performance on planar categories such as \textit{floor} and \textit{wall}, which highlights the effectiveness of our geometry-guided refinement module in enforcing structural consistency aligned with indoor planar layouts. 

The side-by-side visualization comparisons are shown in Fig.~\ref{fig:vis_mono}. Compared to baseline EmbodiedOcc, our method presents better geometric structures in challenging indoor scenes. 
These results collectively validate the effectiveness of our proposed geometry-guided refinement module in enhancing local occupancy prediction across both structured and cluttered indoor regions.

\textbf{Embodied Occupancy Prediction.} 
Following the evaluation protocol established in EmbodiedOcc~\cite{wu2024embodiedocc}, we assess our EmbodiedOcc++ on the challenging task of embodied occupancy prediction. We focus on global scene occupancy prediction after processing all 30 frames in a sequence, evaluating the model's ability to integrate information across multiple views. For baseline SplicingOcc, we follow EmbodiedOcc by using spliced local occupancy predictions from the local occupancy prediction module. 

As shown in Tab.~\ref{tab:online}, EmbodiedOcc++ consistently outperforms both the original EmbodiedOcc~\cite{wu2024embodiedocc} and the SplicingOcc baseline across all evaluation settings. Remarkably, these improvements are achieved without introducing any additional trainable parameters, demonstrating the effectiveness of our geometry-aware and uncertainty-driven techniques within the same architectural backbone. Our method exhibits stronger performance across both structural (e.g., \textit{wall}, \textit{furniture}) and object-centric (e.g., \textit{bed}, \textit{sofa}, \textit{table}) categories, indicating improved spatial consistency and semantic fidelity. These gains stem from the Geometry-guided Refinement Module, which encourages planar alignment, and the Semantic-aware Uncertainty Sampler, which improves update reliability in overlapping regions. Qualitative results in Fig.~\ref{fig:vis_embodied} demonstrate that our approach yields more consistent and structurally precise occupancy predictions, effectively preserving sharp edges and planar boundaries across sequential updates from multiple viewpoints.

\vspace{-2mm}
\subsection{Component Analysis}
% \vspace{-2mm}

\textbf{Component-wise Analysis.} 
To evaluate the individual contributions of the proposed components, we conduct a component-wise ablation study under both the original EmbodiedOcc framework and our improved EmbodiedOcc++ setting. As shown in Tab.\ref{tab:comparison}, we examine the effects of the Geometry-guided Refinement Module and Semantic-aware Uncertainty Sampler. For each experiment, we either retain the original local occupancy checkpoint from EmbodiedOcc~\cite{wu2024embodiedocc} or use the improved checkpoint trained under our EmbodiedOcc++ local prediction framework. 

% \begin{table*}[!h]
%     \vspace{-4mm}
%     \caption{Embodied occupancy prediction performance under different module configurations.}
%     \vspace{-5mm}
%     \small
%     \setlength{\tabcolsep}{0.008\textwidth}
%     \captionsetup{font=scriptsize}
%     \begin{center}
%     \resizebox{0.75\linewidth}{!}{
%     \begin{tabular}{cccc|c|c}
%         \toprule
%         EmbodiedOcc Checkpoint & EmbodiedOcc++ Checkpoint & Geometric Constraints &  Uncertainty Sampler & IoU & mIoU \\
%         \midrule
%         \checkmark & & & & 0.507 & 0.416 \\
%         \checkmark & & \checkmark & & 0.512 & 0.422 \\
%         \checkmark & & & \checkmark & 0.506 & 0.418 \\
%         \checkmark & & \checkmark & \checkmark & 0.500 & 0.424 \\
%         & \checkmark & \checkmark & & 0.528 & 0.433 \\
%         & \checkmark & & \checkmark & 0.527 & 0.431 \\
%         & \checkmark & \checkmark & \checkmark & \textbf{0.529} & \textbf{0.437} \\
%         \bottomrule
%     \end{tabular}}
%     \end{center}
%     \label{tab:comparison}
%     \vspace{-5mm}
% \end{table*}

\begin{table}[!ht]
    \vspace{-3mm}
    \caption{Embodied occupancy prediction performance under different module configurations.}
    \vspace{-3mm}
    \small
    \setlength{\tabcolsep}{0.006\textwidth}
    \centering
    \begin{tabular}{c c | c c | c c}
        \toprule
        \begin{tabular}[c]{@{}c@{}}EmbodiedOcc\\Checkpoint\end{tabular} &
        \begin{tabular}[c]{@{}c@{}}EmbodiedOcc++\\Checkpoint\end{tabular} &
        \begin{tabular}[c]{@{}c@{}}Geometric\\Constraints\end{tabular} &
        \begin{tabular}[c]{@{}c@{}}Uncertainty\\Sampler\end{tabular} &
        IoU & mIoU \\
        \midrule
        \checkmark & & & & 0.507 & 0.416 \\
        \checkmark & & \checkmark & & 0.512 & 0.422 \\
        \checkmark & & & \checkmark & 0.506 & 0.418 \\
        \checkmark & & \checkmark & \checkmark & 0.500 & 0.424 \\
        \midrule
        & \checkmark & \checkmark & & 0.528 & 0.433 \\
        & \checkmark & & \checkmark & 0.527 & 0.431 \\
        & \checkmark & \checkmark & \checkmark & \textbf{0.529} & \textbf{0.437} \\
        \bottomrule
    \end{tabular}
    \label{tab:comparison}
    % \vspace{-5mm}
\end{table}

As presented in Tab.~\ref{tab:comparison}, applying the geometry-guided refinement module based on the EmbodiedOcc checkpoint leads to gains in performance, validating the benefit of introducing geometric regularization. In contrast, using only the uncertainty sampler brings marginal improvements, suggesting that updates are less effective without strong geometric constraints. Interestingly, combining both modules based on the EmbodiedOcc checkpoint introduces minor degradation, likely due to the limited local occupancy perception of EmbodiedOcc. When switching to the EmbodiedOcc++ checkpoint, which provides stronger geometry-aware local predictions, both modules demonstrate clear effectiveness. The geometric refinement improves structural alignment, while the uncertainty-guided updates reduce redundant refinement. 
These findings highlight that the effectiveness of each component depends not only on its own design but also on the ability of the local occupancy prediction.

\begin{table}[!h]
    \vspace{-2mm}
    \caption{
        \textbf{Local occupancy prediction performance with different fusion strategies.}
    }
    \vspace{-4mm}
    \small
    \setlength{\tabcolsep}{0.008\textwidth}
    \captionsetup{font=scriptsize}
    \begin{center}
    \resizebox{0.8\linewidth}{!}{
    \begin{tabular}{c|c|c}
        \toprule
        Fusion Strategy & IoU & mIoU \\
        \midrule
        Surface Curvature based Constraint & 0.551 & 0.473 \\
        Depth-aware Spatial Constraint & 0.548 & 0.476 \\
        Adaptive Constraint Fusion & \textbf{0.557} & \textbf{0.482} \\
        \bottomrule
    \end{tabular}}
    \end{center}
    \label{tab:fusion_strategies}
    \vspace{-4mm}
\end{table}
\textbf{Analysis of the Adaptive Constraint Fusion.}
We evaluate our constraint fusion strategies for local occupancy prediction. Tab.~\ref{tab:fusion_strategies} compares three approaches on the EmbodiedOcc-ScanNet-mini dataset. The Surface Curvature based Constraint uses the curvature map to dynamically adjust normal constraint weights, preserving complex surface details in high-curvature regions. The Depth-aware Spatial Constraint modulates constraint intensity based on point-to-depth distances, showing advantages with planar surfaces. Our Adaptive Constraint Fusion combines these complementary approaches through multiplication, outperforming both standalone methods with improved performance. This fusion applies strong constraints only when both curvature and depth cues agree, preventing over-constraining in uncertain regions. The results demonstrate that different geometric properties provide complementary information, and our approach effectively leverages their strengths for more accurate predictions across diverse scene structures.

\textbf{Analysis of the Uncertainty Point Update.}
We investigate the impact of uncertainty thresholds in our proposed Semantic-aware Uncertainty Sampler on embodied occupancy prediction. 
Tab.~\ref{tab:unc} shows that with a high threshold of 0.7, the model exhibits conservative update behavior. Reducing the threshold to 0.5 improves both IoU and mIoU scores. 
Optimal performance is achieved with a threshold of 0.3, resulting in significantly improved metrics. Additionally, setting the threshold too low diminishes the effect of uncertainty sampling, leading to redundant updates and degraded performance. This validates our approach of using entropy as an update ratio to determine which regions benefit from new observations. Our uncertainty update improves consistency when handling overlapping regions observed from different viewpoints. These results highlight the importance of this mechanism for robust embodied occupancy prediction in sequential observation scenarios. 

\begin{table}[!h]
    \vspace{-2mm}
    \caption{Embodied occupancy prediction performance with different uncertainty thresholds.}
    \vspace{-4mm}
    \small
    \setlength{\tabcolsep}{0.008\textwidth}
    \captionsetup{font=scriptsize}
    \begin{center}
    \resizebox{0.6\linewidth}{!}{
    \begin{tabular}{c|c|c}
        \toprule
        Uncertainty threshold & IoU & mIoU \\
        \midrule
        0.7 & 0.510 & 0.404 \\
        0.5 & 0.522 & 0.423 \\
        0.3 & \textbf{0.529} & \textbf{0.437} \\
        0.1 & 0.521 & 0.427 \\
        \bottomrule
    \end{tabular}}
    \end{center}
    \label{tab:unc}
    \vspace{-4mm}
\end{table}

% \setlength{\intextsep}{3mm}   % 控制表格与上下文的垂直间距（可选）
% \setlength{\columnsep}{3mm}   % 控制表格与正文之间的水平间距
% \begin{wraptable}{r}{0.45\linewidth}
%     \vspace{-3mm}
%     \caption{Embodied occupancy prediction performance with different uncertainty thresholds.}
%     \vspace{-3mm}
%     \small
%     \setlength{\tabcolsep}{0.008\textwidth}
%     \centering
%     \begin{tabular}{c|c|c}
%         \toprule
%         Threshold & IoU & mIoU \\
%         \midrule
%         0.7 & 0.510 & 0.404 \\
%         0.5 & 0.522 & 0.423 \\
%         0.3 & \textbf{0.529} & \textbf{0.437} \\
%         0.1 & 0.521 & 0.427 \\
%         \bottomrule
%     \end{tabular}
%     \label{tab:unc}
%     \vspace{-2mm}
% \end{wraptable}

%%%%%%%%%%%%%%%%%%%%%%%%%%%%%%%%%%%%%%%
% \vspace{-4mm}
\section{Conclusion}
\vspace{-1mm}
In this paper, we present EmbodiedOcc++, the first framework incorporating plane regularization into indoor 3D occupancy prediction. 
We propose three key contributions: (1) a Geometry-guided Refinement Module that constrains Gaussian updates via plane regularization, adaptively enforcing strong constraints only when both curvature and depth cues indicate planar regions, (2) a Semantic-aware Uncertainty Sampler for robust updates in overlapping regions. 
Extensive experiments on the EmbodiedOcc-ScanNet dataset demonstrate that EmbodiedOcc++ consistently outperforms baselines across different settings, with notable improvements in capturing planar structures prevalent in indoor environments. 
Our work highlights the importance of geometric constraints and uncertainty estimation in 3D scene understanding, advancing the capabilities of embodied multimodal and multimedia perception systems operating in complex indoor scenes. 

Although EmbodiedOcc++ effectively handles static indoor environments, it assumes a static world during exploration. Extending EmbodiedOcc++ to dynamic embodied scenarios, where both agents and surrounding objects may move, poses significant challenges and represents an important direction for future research.

\begin{acks}
This work was supported by the National Natural Science Foundation of China (62476011).
\end{acks}

\bibliographystyle{ACM-Reference-Format}
\balance
\bibliography{sample-base}


\begin{thebibliography}{57}

%%% the Web address.

\ifx \showCODEN    \undefined \def \showCODEN     #1{\unskip}     \fi
\ifx \showISBNx    \undefined \def \showISBNx     #1{\unskip}     \fi
\ifx \showISBNxiii \undefined \def \showISBNxiii  #1{\unskip}     \fi
\ifx \showISSN     \undefined \def \showISSN      #1{\unskip}     \fi
\ifx \showLCCN     \undefined \def \showLCCN      #1{\unskip}     \fi
\ifx \shownote     \undefined \def \shownote      #1{#1}          \fi
\ifx \showarticletitle \undefined \def \showarticletitle #1{#1}   \fi
\ifx \showURL      \undefined \def \showURL       {\relax}        \fi
% invisible to TeX
\providecommand\bibfield[2]{#2}
\providecommand\bibinfo[2]{#2}
\providecommand\natexlab[1]{#1}
\providecommand\showeprint[2][]{arXiv:#2}

\bibitem[Bae et~al\mbox{.}(2021)]%
        {bae2021estimating}
\bibfield{author}{\bibinfo{person}{Gwangbin Bae}, \bibinfo{person}{Ignas Budvytis}, {and} \bibinfo{person}{Roberto Cipolla}.} \bibinfo{year}{2021}\natexlab{}.
\newblock \showarticletitle{Estimating and exploiting the aleatoric uncertainty in surface normal estimation}. In \bibinfo{booktitle}{\emph{Proceedings of the IEEE/CVF International Conference on Computer Vision}}. \bibinfo{publisher}{IEEE/CVF}, \bibinfo{address}{Virtual}, \bibinfo{pages}{13137--13146}.
\newblock


\bibitem[Cao and De~Charette(2022)]%
        {cao2022monoscene}
\bibfield{author}{\bibinfo{person}{Anh-Quan Cao} {and} \bibinfo{person}{Raoul De~Charette}.} \bibinfo{year}{2022}\natexlab{}.
\newblock \showarticletitle{Monoscene: Monocular 3d semantic scene completion}. In \bibinfo{booktitle}{\emph{Proceedings of the IEEE/CVF Conference on Computer Vision and Pattern Recognition}}. \bibinfo{publisher}{IEEE/CVF}, \bibinfo{address}{New Orleans, LA, USA}, \bibinfo{pages}{3991--4001}.
\newblock


\bibitem[Chen et~al\mbox{.}(2024)]%
        {chen2024pgsr}
\bibfield{author}{\bibinfo{person}{Danpeng Chen}, \bibinfo{person}{Hai Li}, \bibinfo{person}{Weicai Ye}, \bibinfo{person}{Yifan Wang}, \bibinfo{person}{Weijian Xie}, \bibinfo{person}{Shangjin Zhai}, \bibinfo{person}{Nan Wang}, \bibinfo{person}{Haomin Liu}, \bibinfo{person}{Hujun Bao}, {and} \bibinfo{person}{Guofeng Zhang}.} \bibinfo{year}{2024}\natexlab{}.
\newblock \showarticletitle{Pgsr: Planar-based gaussian splatting for efficient and high-fidelity surface reconstruction}.
\newblock \bibinfo{journal}{\emph{IEEE Transactions on Visualization and Computer Graphics}} (\bibinfo{year}{2024}).
\newblock


\bibitem[Gal and Ghahramani(2016)]%
        {gal2016dropout}
\bibfield{author}{\bibinfo{person}{Yarin Gal} {and} \bibinfo{person}{Zoubin Ghahramani}.} \bibinfo{year}{2016}\natexlab{}.
\newblock \showarticletitle{Dropout as a bayesian approximation: Representing model uncertainty in deep learning}. In \bibinfo{booktitle}{\emph{International Conference on Machine Learning}}. PMLR, \bibinfo{publisher}{PMLR}, \bibinfo{pages}{1050--1059}.
\newblock


\bibitem[Gao et~al\mbox{.}(2023)]%
        {gao2023surfelnerf}
\bibfield{author}{\bibinfo{person}{Yiming Gao}, \bibinfo{person}{Yan-Pei Cao}, {and} \bibinfo{person}{Ying Shan}.} \bibinfo{year}{2023}\natexlab{}.
\newblock \showarticletitle{Surfelnerf: Neural surfel radiance fields for online photorealistic reconstruction of indoor scenes}. In \bibinfo{booktitle}{\emph{Proceedings of the IEEE/CVF Conference on Computer Vision and Pattern Recognition}}. \bibinfo{publisher}{IEEE/CVF}, \bibinfo{address}{Vancouver, Canada}, \bibinfo{pages}{108--118}.
\newblock


\bibitem[Gu{\'e}don and Lepetit(2024)]%
        {guedon2024sugar}
\bibfield{author}{\bibinfo{person}{Antoine Gu{\'e}don} {and} \bibinfo{person}{Vincent Lepetit}.} \bibinfo{year}{2024}\natexlab{}.
\newblock \showarticletitle{Sugar: Surface-aligned gaussian splatting for efficient 3d mesh reconstruction and high-quality mesh rendering}. In \bibinfo{booktitle}{\emph{Proceedings of the IEEE/CVF Conference on Computer Vision and Pattern Recognition}}. \bibinfo{publisher}{IEEE/CVF}, \bibinfo{address}{Vancouver, Canada}, \bibinfo{pages}{5354--5363}.
\newblock


\bibitem[Hou et~al\mbox{.}(2024)]%
        {hou2024fastocc}
\bibfield{author}{\bibinfo{person}{Jiawei Hou}, \bibinfo{person}{Xiaoyan Li}, \bibinfo{person}{Wenhao Guan}, \bibinfo{person}{Gang Zhang}, \bibinfo{person}{Di Feng}, \bibinfo{person}{Yuheng Du}, \bibinfo{person}{Xiangyang Xue}, {and} \bibinfo{person}{Jian Pu}.} \bibinfo{year}{2024}\natexlab{}.
\newblock \showarticletitle{Fastocc: Accelerating 3d occupancy prediction by fusing the 2d bird’s-eye view and perspective view}. In \bibinfo{booktitle}{\emph{2024 IEEE International Conference on Robotics and Automation (ICRA)}}. \bibinfo{publisher}{IEEE}, \bibinfo{address}{Orlando, FL, USA}, \bibinfo{pages}{16425--16431}.
\newblock


\bibitem[Huang et~al\mbox{.}(2024b)]%
        {huang2024textit}
\bibfield{author}{\bibinfo{person}{Nan Huang}, \bibinfo{person}{Xiaobao Wei}, \bibinfo{person}{Wenzhao Zheng}, \bibinfo{person}{Pengju An}, \bibinfo{person}{Ming Lu}, \bibinfo{person}{Wei Zhan}, \bibinfo{person}{Masayoshi Tomizuka}, \bibinfo{person}{Kurt Keutzer}, {and} \bibinfo{person}{Shanghang Zhang}.} \bibinfo{year}{2024}\natexlab{b}.
\newblock \showarticletitle{S3Gaussian: Self-Supervised Street Gaussians for Autonomous Driving}.
\newblock \bibinfo{journal}{\emph{arXiv preprint arXiv:2405.20323}} (\bibinfo{year}{2024}).
\newblock


\bibitem[Huang et~al\mbox{.}(2024a)]%
        {huang2024probabilistic}
\bibfield{author}{\bibinfo{person}{Yuanhui Huang}, \bibinfo{person}{Amonnut Thammatadatrakoon}, \bibinfo{person}{Wenzhao Zheng}, \bibinfo{person}{Yunpeng Zhang}, \bibinfo{person}{Dalong Du}, {and} \bibinfo{person}{Jiwen Lu}.} \bibinfo{year}{2024}\natexlab{a}.
\newblock \showarticletitle{Probabilistic Gaussian Superposition for Efficient 3D Occupancy Prediction}.
\newblock \bibinfo{journal}{\emph{arXiv preprint arXiv:2412.04384}} (\bibinfo{year}{2024}).
\newblock


\bibitem[Huang et~al\mbox{.}(2024c)]%
        {huang2024selfocc}
\bibfield{author}{\bibinfo{person}{Yuanhui Huang}, \bibinfo{person}{Wenzhao Zheng}, \bibinfo{person}{Borui Zhang}, \bibinfo{person}{Jie Zhou}, {and} \bibinfo{person}{Jiwen Lu}.} \bibinfo{year}{2024}\natexlab{c}.
\newblock \showarticletitle{Selfocc: Self-supervised vision-based 3d occupancy prediction}. In \bibinfo{booktitle}{\emph{Proceedings of the IEEE/CVF Conference on Computer Vision and Pattern Recognition}}. \bibinfo{publisher}{IEEE/CVF}, \bibinfo{address}{Vancouver, Canada}, \bibinfo{pages}{19946--19956}.
\newblock


\bibitem[Huang et~al\mbox{.}(2023)]%
        {huang2023tri}
\bibfield{author}{\bibinfo{person}{Yuanhui Huang}, \bibinfo{person}{Wenzhao Zheng}, \bibinfo{person}{Yunpeng Zhang}, \bibinfo{person}{Jie Zhou}, {and} \bibinfo{person}{Jiwen Lu}.} \bibinfo{year}{2023}\natexlab{}.
\newblock \showarticletitle{Tri-perspective view for vision-based 3d semantic occupancy prediction}. In \bibinfo{booktitle}{\emph{Proceedings of the IEEE/CVF Conference on Computer Vision and Pattern Recognition}}. \bibinfo{publisher}{IEEE/CVF}, \bibinfo{address}{Vancouver, Canada}, \bibinfo{pages}{9223--9232}.
\newblock


\bibitem[Huang et~al\mbox{.}(2024d)]%
        {huang2024gaussianformer}
\bibfield{author}{\bibinfo{person}{Yuanhui Huang}, \bibinfo{person}{Wenzhao Zheng}, \bibinfo{person}{Yunpeng Zhang}, \bibinfo{person}{Jie Zhou}, {and} \bibinfo{person}{Jiwen Lu}.} \bibinfo{year}{2024}\natexlab{d}.
\newblock \showarticletitle{Gaussianformer: Scene as gaussians for vision-based 3d semantic occupancy prediction}. In \bibinfo{booktitle}{\emph{European Conference on Computer Vision}}. Springer, \bibinfo{publisher}{Springer}, \bibinfo{address}{Berlin, Germany}, \bibinfo{pages}{376--393}.
\newblock


\bibitem[Humblot-Renaux et~al\mbox{.}(2022)]%
        {humblot2022navigation}
\bibfield{author}{\bibinfo{person}{Galadrielle Humblot-Renaux}, \bibinfo{person}{Letizia Marchegiani}, \bibinfo{person}{Thomas~B Moeslund}, {and} \bibinfo{person}{Rikke Gade}.} \bibinfo{year}{2022}\natexlab{}.
\newblock \showarticletitle{Navigation-oriented scene understanding for robotic autonomy: Learning to segment driveability in egocentric images}.
\newblock \bibinfo{journal}{\emph{IEEE Robotics and Automation Letters}} \bibinfo{volume}{7}, \bibinfo{number}{2} (\bibinfo{year}{2022}), \bibinfo{pages}{2913--2920}.
\newblock


\bibitem[Keetha et~al\mbox{.}(2024)]%
        {keetha2024splatam}
\bibfield{author}{\bibinfo{person}{Nikhil Keetha}, \bibinfo{person}{Jay Karhade}, \bibinfo{person}{Krishna~Murthy Jatavallabhula}, \bibinfo{person}{Gengshan Yang}, \bibinfo{person}{Sebastian Scherer}, \bibinfo{person}{Deva Ramanan}, {and} \bibinfo{person}{Jonathon Luiten}.} \bibinfo{year}{2024}\natexlab{}.
\newblock \showarticletitle{Splatam: Splat track \& map 3d gaussians for dense rgb-d slam}. In \bibinfo{booktitle}{\emph{Proceedings of the IEEE/CVF Conference on Computer Vision and Pattern Recognition}}. \bibinfo{publisher}{IEEE/CVF}, \bibinfo{address}{TBA}, \bibinfo{pages}{21357--21366}.
\newblock


\bibitem[Kerbl et~al\mbox{.}(2023)]%
        {kerbl20233d}
\bibfield{author}{\bibinfo{person}{Bernhard Kerbl}, \bibinfo{person}{Georgios Kopanas}, \bibinfo{person}{Thomas Leimk{\"u}hler}, {and} \bibinfo{person}{George Drettakis}.} \bibinfo{year}{2023}\natexlab{}.
\newblock \showarticletitle{3d gaussian splatting for real-time radiance field rendering.}
\newblock \bibinfo{journal}{\emph{ACM Trans. Graph.}} \bibinfo{volume}{42}, \bibinfo{number}{4} (\bibinfo{year}{2023}), \bibinfo{pages}{139--1}.
\newblock


\bibitem[Kong et~al\mbox{.}(2025)]%
        {kong2025multi}
\bibfield{author}{\bibinfo{person}{Lingdong Kong}, \bibinfo{person}{Xiang Xu}, \bibinfo{person}{Jiawei Ren}, \bibinfo{person}{Wenwei Zhang}, \bibinfo{person}{Liang Pan}, \bibinfo{person}{Kai Chen}, \bibinfo{person}{Wei~Tsang Ooi}, {and} \bibinfo{person}{Ziwei Liu}.} \bibinfo{year}{2025}\natexlab{}.
\newblock \showarticletitle{Multi-modal data-efficient 3d scene understanding for autonomous driving}.
\newblock \bibinfo{journal}{\emph{IEEE Transactions on Pattern Analysis and Machine Intelligence}} \bibinfo{volume}{43}, \bibinfo{number}{11} (\bibinfo{year}{2025}), \bibinfo{pages}{11}.
\newblock


\bibitem[Li et~al\mbox{.}(2022)]%
        {li2022interactive}
\bibfield{author}{\bibinfo{person}{Changyang Li}, \bibinfo{person}{Wanwan Li}, \bibinfo{person}{Haikun Huang}, {and} \bibinfo{person}{Lap-Fai Yu}.} \bibinfo{year}{2022}\natexlab{}.
\newblock \showarticletitle{Interactive augmented reality storytelling guided by scene semantics}.
\newblock \bibinfo{journal}{\emph{ACM Transactions on Graphics (TOG)}} \bibinfo{volume}{41}, \bibinfo{number}{4} (\bibinfo{year}{2022}), \bibinfo{pages}{1--15}.
\newblock


\bibitem[Li et~al\mbox{.}(2023a)]%
        {li2023mseg3d}
\bibfield{author}{\bibinfo{person}{Jiale Li}, \bibinfo{person}{Hang Dai}, \bibinfo{person}{Hao Han}, {and} \bibinfo{person}{Yong Ding}.} \bibinfo{year}{2023}\natexlab{a}.
\newblock \showarticletitle{Mseg3d: Multi-modal 3d semantic segmentation for autonomous driving}. In \bibinfo{booktitle}{\emph{Proceedings of the IEEE/CVF Conference on Computer Vision and Pattern Recognition}}. \bibinfo{publisher}{IEEE/CVF}, \bibinfo{address}{Vancouver, Canada}, \bibinfo{pages}{21694--21704}.
\newblock


\bibitem[Li et~al\mbox{.}(2025)]%
        {li2025sliceocc}
\bibfield{author}{\bibinfo{person}{Jianing Li}, \bibinfo{person}{Ming Lu}, \bibinfo{person}{Hao Wang}, \bibinfo{person}{Chenyang Gu}, \bibinfo{person}{Wenzhao Zheng}, \bibinfo{person}{Li Du}, {and} \bibinfo{person}{Shanghang Zhang}.} \bibinfo{year}{2025}\natexlab{}.
\newblock \showarticletitle{SliceOcc: Indoor 3D Semantic Occupancy Prediction with Vertical Slice Representation}.
\newblock \bibinfo{journal}{\emph{arXiv preprint arXiv:2501.16684}} (\bibinfo{year}{2025}).
\newblock


\bibitem[Li et~al\mbox{.}(2023b)]%
        {li2023neuralangelo}
\bibfield{author}{\bibinfo{person}{Zhaoshuo Li}, \bibinfo{person}{Thomas M{\"u}ller}, \bibinfo{person}{Alex Evans}, \bibinfo{person}{Russell~H Taylor}, \bibinfo{person}{Mathias Unberath}, \bibinfo{person}{Ming-Yu Liu}, {and} \bibinfo{person}{Chen-Hsuan Lin}.} \bibinfo{year}{2023}\natexlab{b}.
\newblock \showarticletitle{Neuralangelo: High-fidelity neural surface reconstruction}. In \bibinfo{booktitle}{\emph{Proceedings of the IEEE/CVF Conference on Computer Vision and Pattern Recognition}}. \bibinfo{publisher}{IEEE/CVF}, \bibinfo{address}{Vancouver, Canada}, \bibinfo{pages}{8456--8465}.
\newblock


\bibitem[Lin et~al\mbox{.}(2024)]%
        {lin2024vastgaussian}
\bibfield{author}{\bibinfo{person}{Jiaqi Lin}, \bibinfo{person}{Zhihao Li}, \bibinfo{person}{Xiao Tang}, \bibinfo{person}{Jianzhuang Liu}, \bibinfo{person}{Shiyong Liu}, \bibinfo{person}{Jiayue Liu}, \bibinfo{person}{Yangdi Lu}, \bibinfo{person}{Xiaofei Wu}, \bibinfo{person}{Songcen Xu}, \bibinfo{person}{Youliang Yan}, {et~al\mbox{.}}} \bibinfo{year}{2024}\natexlab{}.
\newblock \showarticletitle{Vastgaussian: Vast 3d gaussians for large scene reconstruction}. In \bibinfo{booktitle}{\emph{Proceedings of the IEEE/CVF Conference on Computer Vision and Pattern Recognition}}. \bibinfo{publisher}{IEEE/CVF}, \bibinfo{address}{Vancouver, Canada}, \bibinfo{pages}{5166--5175}.
\newblock


\bibitem[Mescheder et~al\mbox{.}(2019)]%
        {mescheder2019occupancy}
\bibfield{author}{\bibinfo{person}{Lars Mescheder}, \bibinfo{person}{Michael Oechsle}, \bibinfo{person}{Michael Niemeyer}, \bibinfo{person}{Sebastian Nowozin}, {and} \bibinfo{person}{Andreas Geiger}.} \bibinfo{year}{2019}\natexlab{}.
\newblock \showarticletitle{Occupancy networks: Learning 3d reconstruction in function space}. In \bibinfo{booktitle}{\emph{Proceedings of the IEEE/CVF Conference on Computer Vision and Pattern Recognition}}. \bibinfo{publisher}{IEEE/CVF}, \bibinfo{address}{Long Beach, CA, USA}, \bibinfo{pages}{4460--4470}.
\newblock


\bibitem[Mildenhall et~al\mbox{.}(2021)]%
        {mildenhall2021nerf}
\bibfield{author}{\bibinfo{person}{Ben Mildenhall}, \bibinfo{person}{Pratul~P Srinivasan}, \bibinfo{person}{Matthew Tancik}, \bibinfo{person}{Jonathan~T Barron}, \bibinfo{person}{Ravi Ramamoorthi}, {and} \bibinfo{person}{Ren Ng}.} \bibinfo{year}{2021}\natexlab{}.
\newblock \showarticletitle{Nerf: Representing scenes as neural radiance fields for view synthesis}.
\newblock \bibinfo{journal}{\emph{Commun. ACM}} \bibinfo{volume}{65}, \bibinfo{number}{1} (\bibinfo{year}{2021}), \bibinfo{pages}{99--106}.
\newblock


\bibitem[Niemeyer et~al\mbox{.}(2019)]%
        {niemeyer2019occupancy}
\bibfield{author}{\bibinfo{person}{Michael Niemeyer}, \bibinfo{person}{Lars Mescheder}, \bibinfo{person}{Michael Oechsle}, {and} \bibinfo{person}{Andreas Geiger}.} \bibinfo{year}{2019}\natexlab{}.
\newblock \showarticletitle{Occupancy flow: 4d reconstruction by learning particle dynamics}. In \bibinfo{booktitle}{\emph{Proceedings of the IEEE/CVF International Conference on Computer Vision}}. \bibinfo{publisher}{IEEE}, \bibinfo{address}{Seoul, South Korea}, \bibinfo{pages}{5379--5389}.
\newblock


\bibitem[Pan et~al\mbox{.}(2024)]%
        {pan2024renderocc}
\bibfield{author}{\bibinfo{person}{Mingjie Pan}, \bibinfo{person}{Jiaming Liu}, \bibinfo{person}{Renrui Zhang}, \bibinfo{person}{Peixiang Huang}, \bibinfo{person}{Xiaoqi Li}, \bibinfo{person}{Hongwei Xie}, \bibinfo{person}{Bing Wang}, \bibinfo{person}{Li Liu}, {and} \bibinfo{person}{Shanghang Zhang}.} \bibinfo{year}{2024}\natexlab{}.
\newblock \showarticletitle{Renderocc: Vision-centric 3d occupancy prediction with 2d rendering supervision}. In \bibinfo{booktitle}{\emph{2024 IEEE International Conference on Robotics and Automation (ICRA)}}. IEEE, \bibinfo{publisher}{IEEE}, \bibinfo{address}{Orlando, FL, USA}, \bibinfo{pages}{12404--12411}.
\newblock


\bibitem[Peng et~al\mbox{.}(2020)]%
        {peng2020convolutional}
\bibfield{author}{\bibinfo{person}{Songyou Peng}, \bibinfo{person}{Michael Niemeyer}, \bibinfo{person}{Lars Mescheder}, \bibinfo{person}{Marc Pollefeys}, {and} \bibinfo{person}{Andreas Geiger}.} \bibinfo{year}{2020}\natexlab{}.
\newblock \showarticletitle{Convolutional occupancy networks}. In \bibinfo{booktitle}{\emph{Computer Vision -- ECCV 2020: 16th European Conference, Glasgow, UK, August 23--28, 2020, Proceedings, Part III}}. Springer, \bibinfo{publisher}{Springer}, \bibinfo{address}{Glasgow, UK}, \bibinfo{pages}{523--540}.
\newblock


\bibitem[Qian et~al\mbox{.}(2024)]%
        {qian2024gaussianavatars}
\bibfield{author}{\bibinfo{person}{Shenhan Qian}, \bibinfo{person}{Tobias Kirschstein}, \bibinfo{person}{Liam Schoneveld}, \bibinfo{person}{Davide Davoli}, \bibinfo{person}{Simon Giebenhain}, {and} \bibinfo{person}{Matthias Nie{\ss}ner}.} \bibinfo{year}{2024}\natexlab{}.
\newblock \showarticletitle{Gaussianavatars: Photorealistic head avatars with rigged 3d gaussians}. In \bibinfo{booktitle}{\emph{Proceedings of the IEEE/CVF Conference on Computer Vision and Pattern Recognition}}. \bibinfo{publisher}{IEEE/CVF}, \bibinfo{address}{Vancouver, Canada}, \bibinfo{pages}{20299--20309}.
\newblock


\bibitem[Sayed et~al\mbox{.}(2022)]%
        {sayed2022simplerecon}
\bibfield{author}{\bibinfo{person}{Mohamed Sayed}, \bibinfo{person}{John Gibson}, \bibinfo{person}{Jamie Watson}, \bibinfo{person}{Victor Prisacariu}, \bibinfo{person}{Michael Firman}, {and} \bibinfo{person}{Cl{\'e}ment Godard}.} \bibinfo{year}{2022}\natexlab{}.
\newblock \showarticletitle{Simplerecon: 3d reconstruction without 3d convolutions}. In \bibinfo{booktitle}{\emph{European Conference on Computer Vision}}. Springer, \bibinfo{publisher}{Springer}, \bibinfo{address}{Tel Aviv, Israel}, \bibinfo{pages}{1--19}.
\newblock


\bibitem[Stier et~al\mbox{.}(2021)]%
        {stier2021vortx}
\bibfield{author}{\bibinfo{person}{Noah Stier}, \bibinfo{person}{Alexander Rich}, \bibinfo{person}{Pradeep Sen}, {and} \bibinfo{person}{Tobias H{\"o}llerer}.} \bibinfo{year}{2021}\natexlab{}.
\newblock \showarticletitle{Vortx: Volumetric 3d reconstruction with transformers for voxelwise view selection and fusion}. In \bibinfo{booktitle}{\emph{2021 International Conference on 3D Vision (3DV)}}. IEEE, \bibinfo{publisher}{IEEE}, \bibinfo{address}{Paris, France}, \bibinfo{pages}{320--330}.
\newblock


\bibitem[Sun et~al\mbox{.}(2021)]%
        {sun2021neuralrecon}
\bibfield{author}{\bibinfo{person}{Jiaming Sun}, \bibinfo{person}{Yiming Xie}, \bibinfo{person}{Linghao Chen}, \bibinfo{person}{Xiaowei Zhou}, {and} \bibinfo{person}{Hujun Bao}.} \bibinfo{year}{2021}\natexlab{}.
\newblock \showarticletitle{Neuralrecon: Real-time coherent 3d reconstruction from monocular video}. In \bibinfo{booktitle}{\emph{Proceedings of the IEEE/CVF Conference on Computer Vision and Pattern Recognition}}. \bibinfo{publisher}{IEEE/CVF}, \bibinfo{address}{Virtual}, \bibinfo{pages}{15598--15607}.
\newblock


\bibitem[Szymanowicz et~al\mbox{.}(2024)]%
        {szymanowicz2024splatter}
\bibfield{author}{\bibinfo{person}{Stanislaw Szymanowicz}, \bibinfo{person}{Chrisitian Rupprecht}, {and} \bibinfo{person}{Andrea Vedaldi}.} \bibinfo{year}{2024}\natexlab{}.
\newblock \showarticletitle{Splatter image: Ultra-fast single-view 3d reconstruction}. In \bibinfo{booktitle}{\emph{Proceedings of the IEEE/CVF Conference on Computer Vision and Pattern Recognition}}. \bibinfo{publisher}{IEEE/CVF}, \bibinfo{address}{Vancouver, Canada}, \bibinfo{pages}{10208--10217}.
\newblock


\bibitem[Tang et~al\mbox{.}(2024)]%
        {tang2024sparseocc}
\bibfield{author}{\bibinfo{person}{Pin Tang}, \bibinfo{person}{Zhongdao Wang}, \bibinfo{person}{Guoqing Wang}, \bibinfo{person}{Jilai Zheng}, \bibinfo{person}{Xiangxuan Ren}, \bibinfo{person}{Bailan Feng}, {and} \bibinfo{person}{Chao Ma}.} \bibinfo{year}{2024}\natexlab{}.
\newblock \showarticletitle{Sparseocc: Rethinking sparse latent representation for vision-based semantic occupancy prediction}. In \bibinfo{booktitle}{\emph{Proceedings of the IEEE/CVF Conference on Computer Vision and Pattern Recognition}}. \bibinfo{publisher}{IEEE/CVF}, \bibinfo{address}{TBA}, \bibinfo{pages}{15035--15044}.
\newblock


\bibitem[Turkulainen et~al\mbox{.}(2024)]%
        {turkulainen2024dn}
\bibfield{author}{\bibinfo{person}{Matias Turkulainen}, \bibinfo{person}{Xuqian Ren}, \bibinfo{person}{Iaroslav Melekhov}, \bibinfo{person}{Otto Seiskari}, \bibinfo{person}{Esa Rahtu}, {and} \bibinfo{person}{Juho Kannala}.} \bibinfo{year}{2024}\natexlab{}.
\newblock \showarticletitle{Dn-splatter: Depth and normal priors for gaussian splatting and meshing}.
\newblock \bibinfo{journal}{\emph{arXiv preprint arXiv:2403.17822}} (\bibinfo{year}{2024}).
\newblock


\bibitem[Wang et~al\mbox{.}(2024a)]%
        {wang2024embodiedscan}
\bibfield{author}{\bibinfo{person}{Tai Wang}, \bibinfo{person}{Xiaohan Mao}, \bibinfo{person}{Chenming Zhu}, \bibinfo{person}{Runsen Xu}, \bibinfo{person}{Ruiyuan Lyu}, \bibinfo{person}{Peisen Li}, \bibinfo{person}{Xiao Chen}, \bibinfo{person}{Wenwei Zhang}, \bibinfo{person}{Kai Chen}, \bibinfo{person}{Tianfan Xue}, {et~al\mbox{.}}} \bibinfo{year}{2024}\natexlab{a}.
\newblock \showarticletitle{Embodiedscan: A holistic multi-modal 3d perception suite towards embodied ai}. In \bibinfo{booktitle}{\emph{Proceedings of the IEEE/CVF Conference on Computer Vision and Pattern Recognition}}. \bibinfo{publisher}{IEEE/CVF}, \bibinfo{address}{Vancouver, Canada}, \bibinfo{pages}{19757--19767}.
\newblock


\bibitem[Wang et~al\mbox{.}(2023)]%
        {wang2023openoccupancy}
\bibfield{author}{\bibinfo{person}{Xiaofeng Wang}, \bibinfo{person}{Zheng Zhu}, \bibinfo{person}{Wenbo Xu}, \bibinfo{person}{Yunpeng Zhang}, \bibinfo{person}{Yi Wei}, \bibinfo{person}{Xu Chi}, \bibinfo{person}{Yun Ye}, \bibinfo{person}{Dalong Du}, \bibinfo{person}{Jiwen Lu}, {and} \bibinfo{person}{Xingang Wang}.} \bibinfo{year}{2023}\natexlab{}.
\newblock \showarticletitle{Openoccupancy: A large scale benchmark for surrounding semantic occupancy perception}. In \bibinfo{booktitle}{\emph{Proceedings of the IEEE/CVF International Conference on Computer Vision}}. \bibinfo{publisher}{IEEE/CVF}, \bibinfo{address}{Vancouver, Canada}, \bibinfo{pages}{17850--17859}.
\newblock


\bibitem[Wang et~al\mbox{.}(2024b)]%
        {wang2024plgs}
\bibfield{author}{\bibinfo{person}{Yu Wang}, \bibinfo{person}{Xiaobao Wei}, \bibinfo{person}{Ming Lu}, {and} \bibinfo{person}{Guoliang Kang}.} \bibinfo{year}{2024}\natexlab{b}.
\newblock \showarticletitle{PLGS: Robust Panoptic Lifting with 3D Gaussian Splatting}.
\newblock \bibinfo{journal}{\emph{arXiv preprint arXiv:2410.17505}} (\bibinfo{year}{2024}).
\newblock


\bibitem[Wei et~al\mbox{.}(2024a)]%
        {wei2024medsam}
\bibfield{author}{\bibinfo{person}{Xiaobao Wei}, \bibinfo{person}{Jiajun Cao}, \bibinfo{person}{Yizhu Jin}, \bibinfo{person}{Ming Lu}, \bibinfo{person}{Guangyu Wang}, {and} \bibinfo{person}{Shanghang Zhang}.} \bibinfo{year}{2024}\natexlab{a}.
\newblock \showarticletitle{I-medsam: Implicit medical image segmentation with segment anything}. In \bibinfo{booktitle}{\emph{European Conference on Computer Vision}}. Springer, \bibinfo{publisher}{Springer}, \bibinfo{address}{Munich, Germany}, \bibinfo{pages}{90--107}.
\newblock


\bibitem[Wei et~al\mbox{.}(2024b)]%
        {wei2024gazegaussian}
\bibfield{author}{\bibinfo{person}{Xiaobao Wei}, \bibinfo{person}{Peng Chen}, \bibinfo{person}{Guangyu Li}, \bibinfo{person}{Ming Lu}, \bibinfo{person}{Hui Chen}, {and} \bibinfo{person}{Feng Tian}.} \bibinfo{year}{2024}\natexlab{b}.
\newblock \showarticletitle{GazeGaussian: High-Fidelity Gaze Redirection with 3D Gaussian Splatting}.
\newblock \bibinfo{journal}{\emph{arXiv preprint arXiv:2411.12981}} (\bibinfo{year}{2024}).
\newblock


\bibitem[Wei et~al\mbox{.}(2024c)]%
        {wei2024graphavatar}
\bibfield{author}{\bibinfo{person}{Xiaobao Wei}, \bibinfo{person}{Peng Chen}, \bibinfo{person}{Ming Lu}, \bibinfo{person}{Hui Chen}, {and} \bibinfo{person}{Feng Tian}.} \bibinfo{year}{2024}\natexlab{c}.
\newblock \showarticletitle{GraphAvatar: Compact Head Avatars with GNN-Generated 3D Gaussians}.
\newblock \bibinfo{journal}{\emph{arXiv preprint arXiv:2412.13983}} (\bibinfo{year}{2024}).
\newblock


\bibitem[Wei et~al\mbox{.}(2024d)]%
        {wei2024emd}
\bibfield{author}{\bibinfo{person}{Xiaobao Wei}, \bibinfo{person}{Qingpo Wuwu}, \bibinfo{person}{Zhongyu Zhao}, \bibinfo{person}{Zhuangzhe Wu}, \bibinfo{person}{Nan Huang}, \bibinfo{person}{Ming Lu}, \bibinfo{person}{Ningning Ma}, {and} \bibinfo{person}{Shanghang Zhang}.} \bibinfo{year}{2024}\natexlab{d}.
\newblock \showarticletitle{EMD: Explicit Motion Modeling for High-Quality Street Gaussian Splatting}.
\newblock \bibinfo{journal}{\emph{arXiv preprint arXiv:2411.15582}} (\bibinfo{year}{2024}).
\newblock


\bibitem[Wei et~al\mbox{.}(2024e)]%
        {wei2024nto3d}
\bibfield{author}{\bibinfo{person}{Xiaobao Wei}, \bibinfo{person}{Renrui Zhang}, \bibinfo{person}{Jiarui Wu}, \bibinfo{person}{Jiaming Liu}, \bibinfo{person}{Ming Lu}, \bibinfo{person}{Yandong Guo}, {and} \bibinfo{person}{Shanghang Zhang}.} \bibinfo{year}{2024}\natexlab{e}.
\newblock \showarticletitle{Nto3d: Neural target object 3d reconstruction with segment anything}. In \bibinfo{booktitle}{\emph{Proceedings of the IEEE/CVF Conference on Computer Vision and Pattern Recognition}}. \bibinfo{publisher}{IEEE/CVF}, \bibinfo{address}{Vancouver, Canada}, \bibinfo{pages}{20352--20362}.
\newblock


\bibitem[Wei et~al\mbox{.}(2025)]%
        {wei2025omniindoor3d}
\bibfield{author}{\bibinfo{person}{Xiaobao Wei}, \bibinfo{person}{Xiaoan Zhang}, \bibinfo{person}{Hao Wang}, \bibinfo{person}{Qingpo Wuwu}, \bibinfo{person}{Ming Lu}, \bibinfo{person}{Wenzhao Zheng}, {and} \bibinfo{person}{Shanghang Zhang}.} \bibinfo{year}{2025}\natexlab{}.
\newblock \showarticletitle{OmniIndoor3D: Comprehensive Indoor 3D Reconstruction}.
\newblock \bibinfo{journal}{\emph{arXiv preprint arXiv:2505.20610}} (\bibinfo{year}{2025}).
\newblock


\bibitem[Wei et~al\mbox{.}(2023)]%
        {wei2023surroundocc}
\bibfield{author}{\bibinfo{person}{Yi Wei}, \bibinfo{person}{Linqing Zhao}, \bibinfo{person}{Wenzhao Zheng}, \bibinfo{person}{Zheng Zhu}, \bibinfo{person}{Jie Zhou}, {and} \bibinfo{person}{Jiwen Lu}.} \bibinfo{year}{2023}\natexlab{}.
\newblock \showarticletitle{Surroundocc: Multi-camera 3d occupancy prediction for autonomous driving}. In \bibinfo{booktitle}{\emph{Proceedings of the IEEE/CVF International Conference on Computer Vision}}. \bibinfo{publisher}{IEEE/CVF}, \bibinfo{address}{Vancouver, Canada}, \bibinfo{pages}{21729--21740}.
\newblock


\bibitem[Wu et~al\mbox{.}(2024)]%
        {wu2024embodiedocc}
\bibfield{author}{\bibinfo{person}{Yuqi Wu}, \bibinfo{person}{Wenzhao Zheng}, \bibinfo{person}{Sicheng Zuo}, \bibinfo{person}{Yuanhui Huang}, \bibinfo{person}{Jie Zhou}, {and} \bibinfo{person}{Jiwen Lu}.} \bibinfo{year}{2024}\natexlab{}.
\newblock \showarticletitle{Embodiedocc: Embodied 3d occupancy prediction for vision-based online scene understanding}.
\newblock \bibinfo{journal}{\emph{arXiv preprint arXiv:2412.04380}} (\bibinfo{year}{2024}).
\newblock


\bibitem[Yan et~al\mbox{.}(2024)]%
        {yan2024gs}
\bibfield{author}{\bibinfo{person}{Chi Yan}, \bibinfo{person}{Delin Qu}, \bibinfo{person}{Dan Xu}, \bibinfo{person}{Bin Zhao}, \bibinfo{person}{Zhigang Wang}, \bibinfo{person}{Dong Wang}, {and} \bibinfo{person}{Xuelong Li}.} \bibinfo{year}{2024}\natexlab{}.
\newblock \showarticletitle{Gs-slam: Dense visual slam with 3d gaussian splatting}. In \bibinfo{booktitle}{\emph{Proceedings of the IEEE/CVF Conference on Computer Vision and Pattern Recognition}}. \bibinfo{publisher}{IEEE/CVF}, \bibinfo{address}{TBA}, \bibinfo{pages}{19595--19604}.
\newblock


\bibitem[Yang et~al\mbox{.}(2024)]%
        {yang2024depth}
\bibfield{author}{\bibinfo{person}{Lihe Yang}, \bibinfo{person}{Bingyi Kang}, \bibinfo{person}{Zilong Huang}, \bibinfo{person}{Zhen Zhao}, \bibinfo{person}{Xiaogang Xu}, \bibinfo{person}{Jiashi Feng}, {and} \bibinfo{person}{Hengshuang Zhao}.} \bibinfo{year}{2024}\natexlab{}.
\newblock \showarticletitle{Depth anything v2}.
\newblock \bibinfo{journal}{\emph{Advances in Neural Information Processing Systems}}  \bibinfo{volume}{37} (\bibinfo{year}{2024}), \bibinfo{pages}{21875--21911}.
\newblock


\bibitem[Yao et~al\mbox{.}(2023)]%
        {yao2023ndc}
\bibfield{author}{\bibinfo{person}{Jiawei Yao}, \bibinfo{person}{Chuming Li}, \bibinfo{person}{Keqiang Sun}, \bibinfo{person}{Yingjie Cai}, \bibinfo{person}{Hao Li}, \bibinfo{person}{Wanli Ouyang}, {and} \bibinfo{person}{Hongsheng Li}.} \bibinfo{year}{2023}\natexlab{}.
\newblock \showarticletitle{Ndc-scene: Boost monocular 3d semantic scene completion in normalized device coordinates space}. In \bibinfo{booktitle}{\emph{2023 IEEE/CVF International Conference on Computer Vision (ICCV)}}. IEEE Computer Society, \bibinfo{publisher}{IEEE}, \bibinfo{address}{Paris, France}, \bibinfo{pages}{9421--9431}.
\newblock


\bibitem[Yariv et~al\mbox{.}(2021)]%
        {yariv2021volume}
\bibfield{author}{\bibinfo{person}{Lior Yariv}, \bibinfo{person}{Jiatao Gu}, \bibinfo{person}{Yoni Kasten}, {and} \bibinfo{person}{Yaron Lipman}.} \bibinfo{year}{2021}\natexlab{}.
\newblock \showarticletitle{Volume rendering of neural implicit surfaces}.
\newblock \bibinfo{journal}{\emph{Advances in Neural Information Processing Systems}}  \bibinfo{volume}{34} (\bibinfo{year}{2021}), \bibinfo{pages}{4805--4815}.
\newblock


\bibitem[Yu et~al\mbox{.}(2024b)]%
        {yu2024monocular}
\bibfield{author}{\bibinfo{person}{Hongxiao Yu}, \bibinfo{person}{Yuqi Wang}, \bibinfo{person}{Yuntao Chen}, {and} \bibinfo{person}{Zhaoxiang Zhang}.} \bibinfo{year}{2024}\natexlab{b}.
\newblock \showarticletitle{Monocular occupancy prediction for scalable indoor scenes}. In \bibinfo{booktitle}{\emph{European Conference on Computer Vision}}. Springer, \bibinfo{publisher}{Springer}, \bibinfo{address}{Berlin, Germany}, \bibinfo{pages}{38--54}.
\newblock


\bibitem[Yu et~al\mbox{.}(2024a)]%
        {yu2024gsdf}
\bibfield{author}{\bibinfo{person}{Mulin Yu}, \bibinfo{person}{Tao Lu}, \bibinfo{person}{Linning Xu}, \bibinfo{person}{Lihan Jiang}, \bibinfo{person}{Yuanbo Xiangli}, {and} \bibinfo{person}{Bo Dai}.} \bibinfo{year}{2024}\natexlab{a}.
\newblock \showarticletitle{Gsdf: 3dgs meets sdf for improved rendering and reconstruction}.
\newblock \bibinfo{journal}{\emph{arXiv preprint arXiv:2403.16964}} \bibinfo{volume}{1}, \bibinfo{number}{1} (\bibinfo{year}{2024}), \bibinfo{pages}{1--12}.
\newblock


\bibitem[Yu et~al\mbox{.}(2022)]%
        {yu2022monosdf}
\bibfield{author}{\bibinfo{person}{Zehao Yu}, \bibinfo{person}{Songyou Peng}, \bibinfo{person}{Michael Niemeyer}, \bibinfo{person}{Torsten Sattler}, {and} \bibinfo{person}{Andreas Geiger}.} \bibinfo{year}{2022}\natexlab{}.
\newblock \showarticletitle{Monosdf: Exploring monocular geometric cues for neural implicit surface reconstruction}.
\newblock \bibinfo{journal}{\emph{Advances in Neural Information Processing Systems}}  \bibinfo{volume}{35} (\bibinfo{year}{2022}), \bibinfo{pages}{25018--25032}.
\newblock


\bibitem[Zhang et~al\mbox{.}(2024)]%
        {zhang2024rade}
\bibfield{author}{\bibinfo{person}{Baowen Zhang}, \bibinfo{person}{Chuan Fang}, \bibinfo{person}{Rakesh Shrestha}, \bibinfo{person}{Yixun Liang}, \bibinfo{person}{Xiaoxiao Long}, {and} \bibinfo{person}{Ping Tan}.} \bibinfo{year}{2024}\natexlab{}.
\newblock \showarticletitle{Rade-gs: Rasterizing depth in gaussian splatting}.
\newblock \bibinfo{journal}{\emph{arXiv preprint arXiv:2406.01467}} (\bibinfo{year}{2024}).
\newblock


\bibitem[Zhang et~al\mbox{.}(2022)]%
        {zhang2022outdoor}
\bibfield{author}{\bibinfo{person}{Fu-sheng Zhang}, \bibinfo{person}{Dong-yuan Ge}, \bibinfo{person}{Jun Song}, {and} \bibinfo{person}{Wen-jiang Xiang}.} \bibinfo{year}{2022}\natexlab{}.
\newblock \showarticletitle{Outdoor scene understanding of mobile robot via multi-sensor information fusion}.
\newblock \bibinfo{journal}{\emph{Journal of Industrial Information Integration}}  \bibinfo{volume}{30} (\bibinfo{year}{2022}), \bibinfo{pages}{100392}.
\newblock


\bibitem[Zhang et~al\mbox{.}(2023)]%
        {zhang2023occformer}
\bibfield{author}{\bibinfo{person}{Yunpeng Zhang}, \bibinfo{person}{Zheng Zhu}, {and} \bibinfo{person}{Dalong Du}.} \bibinfo{year}{2023}\natexlab{}.
\newblock \showarticletitle{Occformer: Dual-path transformer for vision-based 3d semantic occupancy prediction}. In \bibinfo{booktitle}{\emph{Proceedings of the IEEE/CVF International Conference on Computer Vision}}. \bibinfo{publisher}{IEEE/CVF}, \bibinfo{address}{Vancouver, Canada}, \bibinfo{pages}{9433--9443}.
\newblock


\bibitem[Zhao et~al\mbox{.}(2024)]%
        {zhao2024hybridocc}
\bibfield{author}{\bibinfo{person}{Xiao Zhao}, \bibinfo{person}{Bo Chen}, \bibinfo{person}{Mingyang Sun}, \bibinfo{person}{Dingkang Yang}, \bibinfo{person}{Youxing Wang}, \bibinfo{person}{Xukun Zhang}, \bibinfo{person}{Mingcheng Li}, \bibinfo{person}{Dongliang Kou}, \bibinfo{person}{Xiaoyi Wei}, {and} \bibinfo{person}{Lihua Zhang}.} \bibinfo{year}{2024}\natexlab{}.
\newblock \showarticletitle{Hybridocc: Nerf enhanced transformer-based multi-camera 3d occupancy prediction}.
\newblock \bibinfo{journal}{\emph{IEEE Robotics and Automation Letters}} \bibinfo{volume}{9}, \bibinfo{number}{3} (\bibinfo{year}{2024}), \bibinfo{pages}{1234--1245}.
\newblock


\bibitem[Zhu et~al\mbox{.}(2022)]%
        {zhu2022nice}
\bibfield{author}{\bibinfo{person}{Zihan Zhu}, \bibinfo{person}{Songyou Peng}, \bibinfo{person}{Viktor Larsson}, \bibinfo{person}{Weiwei Xu}, \bibinfo{person}{Hujun Bao}, \bibinfo{person}{Zhaopeng Cui}, \bibinfo{person}{Martin~R Oswald}, {and} \bibinfo{person}{Marc Pollefeys}.} \bibinfo{year}{2022}\natexlab{}.
\newblock \showarticletitle{Nice-slam: Neural implicit scalable encoding for slam}. In \bibinfo{booktitle}{\emph{Proceedings of the IEEE/CVF Conference on Computer Vision and Pattern Recognition}}. \bibinfo{publisher}{IEEE/CVF}, \bibinfo{address}{New Orleans, LA, USA}, \bibinfo{pages}{12786--12796}.
\newblock


\bibitem[Zuo et~al\mbox{.}(2023)]%
        {zuo2023pointocc}
\bibfield{author}{\bibinfo{person}{Sicheng Zuo}, \bibinfo{person}{Wenzhao Zheng}, \bibinfo{person}{Yuanhui Huang}, \bibinfo{person}{Jie Zhou}, {and} \bibinfo{person}{Jiwen Lu}.} \bibinfo{year}{2023}\natexlab{}.
\newblock \showarticletitle{Pointocc: Cylindrical tri-perspective view for point-based 3d semantic occupancy prediction}.
\newblock \bibinfo{journal}{\emph{arXiv preprint arXiv:2308.16896}} \bibinfo{volume}{2}, \bibinfo{number}{1} (\bibinfo{year}{2023}), \bibinfo{pages}{101--112}.
\newblock


\end{thebibliography}

\clearpage
\appendix

\section{Overview}
The supplementary material includes the subsequent components. 
% We provide additional training details (Appendix A) and more experimental visualization, along with the per-scene quantitative and qualitative results of our method on all datasets (Appendix B). 
% For additional visual comparisons, we highly encourage you to visit our anonymous website at \href{https://emdgaussian.github.io/}{https://emdgaussian.github.io/}, where we showcase more side-by-side visualization results.
\begin{itemize}
    \item Data Preprocessing
    \item Additional Visualization Results
    \begin{itemize}
        \item[--] Local Occupancy Prediction Visualization
        \item[--] Embodied Occupancy Prediction Visualization
    \end{itemize}
    \item Fusion Strategies for Normal Constraint Weights
    \item Impact of Semantic-aware Uncertainty Module on Computational Efficiency
\end{itemize}

\section{Data Preprocessing}
In our experiments, we conduct additional preprocessing on the EmbodiedOcc-ScanNet dataset~\cite{wu2024embodiedocc}. To improve training efficiency, we utilized a pre-trained model~\cite{bae2021estimating} to precompute surface normals and curvature values for each image in the dataset. Specifically, we run inference using the pre-trained normal model on all images in the EmbodiedOcc-ScanNet dataset and save the generated normal maps and their corresponding confidence maps as supplementary data. This precomputation strategy significantly reduces the computational burden during training by avoiding the repeated execution of the normal estimation model during each training iteration, which would otherwise slow down the training process considerably. We still employ the same pre-trained normal estimation model for real-time inference during the inference phase to maintain consistency between training and testing pipelines. This approach ensures both stable model performance and improved training efficiency. The preprocessed data consists of the original RGB images, precomputed normal maps, and their corresponding confidence maps, which collectively serve as input to our model for the subsequent occupancy prediction task.

\begin{figure}[!bp]
  \includegraphics[width=0.45\textwidth]{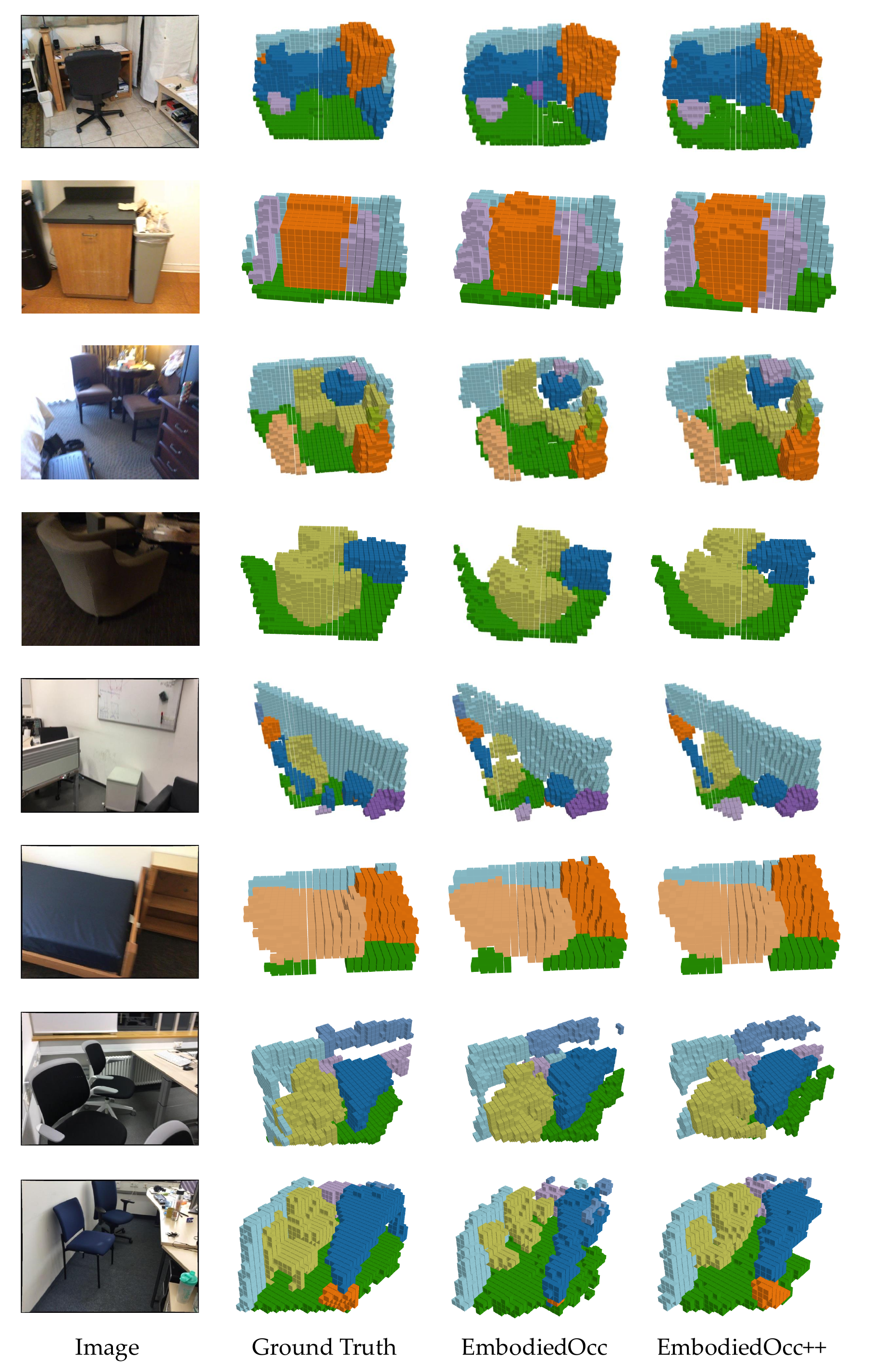}
  \vspace{-3mm}
  \caption{Visualization of our local occupancy prediction. The images show how our model effectively constructs occupancy estimates from single monocular views, preserving spatial relationships within the camera frustum.}
  % \Description{2.}
  % \vspace{-8mm}
  \label{fig:app_mono}
\end{figure}

\section{Additional Visualization Results}
This section provides supplementary visualizations that further illustrate the performance of our approach on both experimental tasks described in the main text. These visualizations offer additional insights beyond what could be accommodated in the primary sections of the paper.

We also include a comparison video in ``demo.mp4"", which contains embodied occupancy prediction comparison between the EmbodiedOcc and ours method.

\subsection{Local Occupancy Prediction Visualization}
The local occupancy prediction task involves using single monocular images to predict occupancy within the camera's frustum. Fig.~\ref{fig:app_mono} demonstrates our model's capability to accurately estimate spatial occupancy from individual frames. The visualization highlights how our approach effectively captures the geometric structure of the scene from limited visual information, preserving both proximal detail and distant elements within the camera's field of view.

\begin{figure*}[!ht]
  \includegraphics[width=\textwidth]{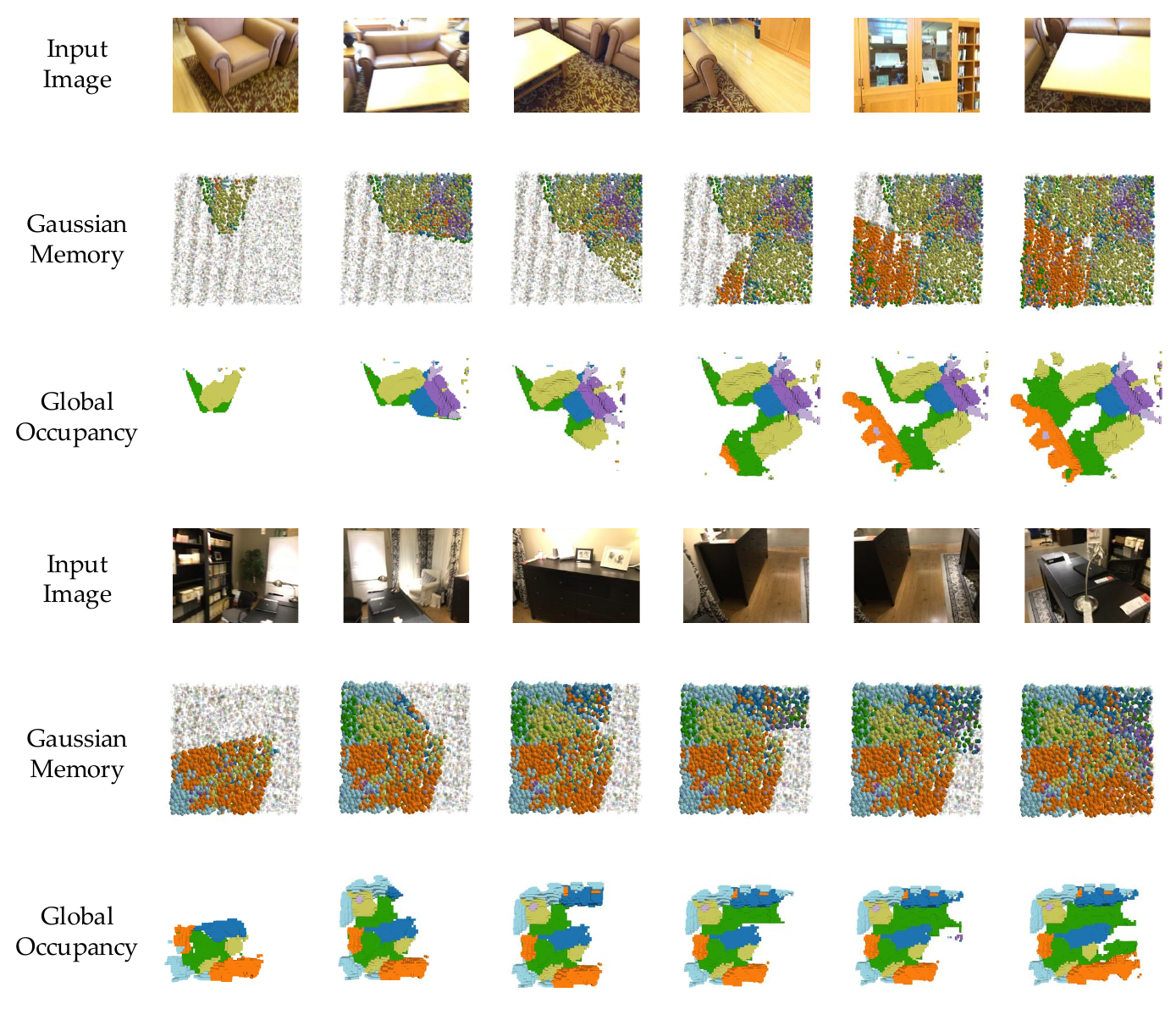}
  % \vspace{-4mm}
  \caption{Visualization of our embodied occupancy prediction. This figure demonstrates the online integration of sequential visual inputs, showing how our model progressively refines its spatial understanding as new observations become available. The temporal sequence illustrates the accumulation of occupancy information across multiple frames.}
  % \vspace{-4mm}
  % \Description{2.}
  \label{fig:app_online}
\end{figure*}

Notably, EmbodiedOcc++ demonstrates robust performance even in challenging conditions such as low-texture surfaces, complex geometries, and varied lighting scenarios. The visualizations reveal how our model successfully distinguishes between solid objects and free space, preserving fine-grained details like furniture edges, wall contours, and small objects. We also observe that the model appropriately captures uncertainty in ambiguous regions, such as object boundaries and partially occluded areas, which aligns with perceptual principles of human depth estimation from monocular cues.

\subsection{Embodied Occupancy Prediction Visualization}
For the more challenging embodied occupancy prediction task, our system processes sequential visual inputs to continuously update occupancy estimates in an online manner. Fig.~\ref{fig:app_online} illustrates this process, showing how our approach integrates information across multiple frames to build and refine a comprehensive spatial understanding. The visualization demonstrates the progressive improvement in occupancy prediction quality as the system accumulates visual evidence over time, highlighting the adaptability of our approach to dynamic environments.

The visualization highlights how our method handles occlusions and view-dependent effects. As the camera navigates through the environment, previously occluded regions become visible and are rapidly incorporated into the spatial representation. Concurrently, the system maintains consistency in regions that are no longer visible but have been previously observed with high confidence. This balance between integration of new information and preservation of established knowledge is crucial for stable and accurate embodied mapping applications.

\definecolor{ceiling}{RGB}{214,  38, 40}   %
\definecolor{floor}{RGB}{43, 160, 4}     %
\definecolor{wall}{RGB}{158, 216, 229}  %
\definecolor{window}{RGB}{114, 158, 206}  %
\definecolor{chair}{RGB}{204, 204, 91}   %
\definecolor{bed}{RGB}{255, 186, 119}  %
\definecolor{sofa}{RGB}{147, 102, 188}  %
\definecolor{table}{RGB}{30, 119, 181}   %
\definecolor{tvs}{RGB}{160, 188, 33}   %
\definecolor{furniture}{RGB}{255, 127, 12}  %
\definecolor{objects}{RGB}{196, 175, 214} %

\begin{table*}[!t]
		\caption{
        \textbf{Comparison of different fusion strategies for integrating depth-aware spatial constraint and surface curvature based constraint on the Occ-ScanNet-mini datasets.}
        }
        \vspace{-4mm}
		\small
		\setlength{\tabcolsep}{0.008\textwidth}
		\captionsetup{font=scriptsize}
            \begin{center}
            \resizebox{1.0\linewidth}{!}{
		\begin{tabular}{l|c|c|c c c c c c c c c c c|c}
			\toprule
			Dataset
			& Method
			& {IoU}
			& \rotatebox{90}{\parbox{1.5cm}{\textcolor{ceiling}{$\blacksquare$} ceiling}} 
			& \rotatebox{90}{\textcolor{floor}{$\blacksquare$} floor}
			& \rotatebox{90}{\textcolor{wall}{$\blacksquare$} wall} 
			& \rotatebox{90}{\textcolor{window}{$\blacksquare$} window} 
			& \rotatebox{90}{\textcolor{chair}{$\blacksquare$} chair} 
			& \rotatebox{90}{\textcolor{bed}{$\blacksquare$} bed} 
			& \rotatebox{90}{\textcolor{sofa}{$\blacksquare$} sofa} 
			& \rotatebox{90}{\textcolor{table}{$\blacksquare$} table} 
			& \rotatebox{90}{\textcolor{tvs}{$\blacksquare$} tvs} 
			& \rotatebox{90}{\textcolor{furniture}{$\blacksquare$} furniture} 
			& \rotatebox{90}{\textcolor{objects}{$\blacksquare$} objects} 
			& mIoU\\
			\midrule
			\multirow{8}{*}{Occ-ScanNet-mini} 
			& Only use Kappa & 0.558 & 0.242 & 0.515 & 0.432 & 0.395 & 0.425 & 0.636 & 0.630 & 0.507 & 0.344 & 0.600 & 0.484 & 0.473 \\
            & Only use Depth & \textbf{0.565} & 0.180 & \textbf{0.516} & \textbf{0.446} & \textbf{0.402} & \textbf{0.441} & 0.650 & 0.636 & 0.510 & 0.364 & \textbf{0.606} & 0.488 & 0.476 \\
            & Weighted sum & 0.534 & 0.219 & 0.483 & 0.427 & 0.362 & 0.418 & 0.622 & 0.605 & 0.475 & 0.306 & 0.575 & 0.463 & 0.450 \\
            & Confidence based & 0.521 & 0.201 & 0.464 & 0.409 & 0.318 & 0.413 & 0.614 & 0.603 & 0.457 & 0.275 & 0.568 & 0.460 & 0.435 \\
            & Max constraints & 0.534 & 0.214 & 0.481 & 0.423 & 0.363 & 0.416 & 0.621 & 0.611 & 0.468 & 0.324 & 0.577 & 0.470 & 0.452 \\
            & Min constraints & 0.533 & 0.233 & 0.480 & 0.418 & 0.360 & 0.415 & 0.625 & 0.604 & 0.474 & 0.318 & 0.578 & 0.470 & 0.452 \\
            & Adaptive & 0.559 & \textbf{0.264} & 0.502 & 0.444 & 0.400 & 0.437 & 0.649 & \textbf{0.641} & 0.502 & 0.365 & 0.603 & 0.470 & 0.479 \\
            & Region Adaptive & 0.539 & 0.259 & 0.489 & 0.425 & 0.382 & 0.422 & 0.626 & 0.622 & 0.480 & 0.303 & 0.581 & 0.472 & 0.460 \\
            & Product (ours) & 0.557 & 0.233 & 0.510 & 0.428 & 0.393 & 0.435 & \textbf{0.656} & 0.640 & \textbf{0.507} & \textbf{0.407} & 0.603 & \textbf{0.489} & \textbf{0.482} \\
		\bottomrule
		\end{tabular}}
        \end{center}
        \vspace{-2mm}
		\label{tab:app_fusion_strategy}
 \end{table*}
\begin{table*}[t]
\caption{\textbf{Comparison of Gaussian updates per frame between EmbodiedOcc and EmbodiedOcc++.} Our method demonstrates more efficient Gaussian allocation and scene representation as frames are progressively integrated.}
% \vspace{-4mm}
\centering
\resizebox{0.9\linewidth}{!}{
\begin{tabular}{c|c|l|ccccccc|c|c}
\toprule
\multirow{2}{*}{Scene} & Initial & \multirow{2}{*}{Method} & \multicolumn{7}{c|}{Number of Updated Gaussian Points at Frame} & \multirow{2}{*}{Total} & \multirow{2}{*}{Avg/Frame} \\
& Total & & 1 & 5 & 10 & 15 & 20 & 25 & 30 & & \\
\midrule
\multirow{2}{*}{scene0089\_00} & \multirow{2}{*}{8,400} & EmbodiedOcc & 954 & 756 & 923 & 1173 & 283 & 427 & 759 & 22,835 & 761 \\
& & EmbodiedOcc++ (Ours) & 954 & \textbf{647} & \textbf{703} & \textbf{906} & \textbf{182} & \textbf{323} & \textbf{625} & \textbf{17,911} & \textbf{597} \\
\bottomrule
\end{tabular}
}
% \vspace{-2mm}
\label{tab:point_growth}
\end{table*}
\section{Fusion Strategies for Normal Constraint Weights}
In our implementation, we explore various fusion strategies to combine Depth-aware Spatial Constraint and Surface Curvature based Constraint weights. The weighted sum strategy linearly combines the two information sources with predefined weights, offering direct control over their relative importance. The adaptive strategy dynamically adjusts the weighting based on the reliability of depth information, relying more heavily on depth constraints when available and falling back to curvature guidance when depth information is uncertain. The confidence-based strategy modulates weights based on an estimated confidence measure derived from depth weights, reducing depth influence when uncertain. The max constraint strategy takes the maximum value between weights, representing a conservative approach that prioritizes geometric accuracy, while the min constraint strategy takes the minimum value, allowing more freedom in surface reconstruction. The region adaptive strategy dynamically adjusts fusion weights based on local surface properties, applying different weightings in high-curvature versus low-curvature regions. The product strategy multiplies the depth and curvature weights, resembling a logical "AND" operation where the constraint is strong only when both information sources agree. As shown in Tab.~\ref{tab:app_fusion_strategy}, the product fusion strategy achieves the best overall performance with superior mIoU scores across most object categories, effectively leveraging the strengths of both information sources while mitigating their individual weaknesses.

\section{Impact of Semantic-aware Uncertainty Module on Computational Efficiency}
The Semantic-aware Uncertainty Module introduced in our framework significantly improves computational efficiency during progressive scene reconstruction. As demonstrated in Tab.~\ref{tab:point_growth}, our approach reduces the average number of Gaussian updates required during scene reconstruction from 761 updates per frame in the baseline EmbodiedOcc~\cite{wu2024embodiedocc} to 597 updates per frame in our EmbodiedOcc++, constituting a 21.5\% reduction in computational workload. This improvement can be attributed to the uncertainty-guided update mechanism that selectively processes Gaussians based on their semantic reliability. By establishing an uncertainty threshold $\tau_{unc}$ below which points are considered sufficiently reliable and excluded from updates, our method avoids unnecessary refinement of already well-established scene elements.

As shown in Tab.~\ref{tab:point_growth}, both methods initially update a similar number of points (954 points at frame 1), but our approach consistently requires fewer updates in subsequent frames. This advantage becomes particularly pronounced in later frames (e.g., at frame 20, only 182 points require updates in our method compared to 283 in the baseline) as more scene elements stabilize with accumulated observations. The data across all sampled frames demonstrates that our uncertainty-guided approach progressively reduces computational load as scene understanding improves. The reduction in update operations is especially beneficial in regions of the scene that are repeatedly observed from multiple viewpoints. In the baseline approach, these overlapping regions would trigger redundant updates in every frame, despite having already converged to a stable representation. Our uncertainty-guided approach intelligently determines which points have reached sufficient stability and can be excluded from further processing. Importantly, this computational efficiency is achieved without sacrificing reconstruction quality. As shown in our main experimental results, EmbodiedOcc++ maintains or improves performance across all evaluation metrics compared to the baseline. This demonstrates that our selective update strategy effectively identifies which Gaussians genuinely require refinement, focusing computational resources where they are most needed.

\begin{table}[t]
\caption{\textbf{Inference time comparison for the confidence refinement module in EmbodiedOcc and EmbodiedOcc++.} Our semantic-aware refinement layer requires additional but reasonable computational resources while enabling more intelligent point update decisions.}
% \vspace{-4mm}
\centering
\resizebox{1.0\linewidth}{!}{
\begin{tabular}{l|c|c|c}
\toprule
Method & Confidence Refinement Time (ms) & Overhead (ms) & Overhead (\%) \\
\midrule
EmbodiedOcc & 1.60 & - & - \\
EmbodiedOcc++ (Ours) & 2.25 & +0.65 & +40.6\% \\
\bottomrule
\end{tabular}
}
% \vspace{-2mm}
\label{tab:inference_time}
\end{table}
While our semantic-aware uncertainty module introduces a slight computational overhead in the confidence refinement stage, this trade-off is well justified by the overall computational savings and performance improvements. As shown in Tab.~\ref{tab:inference_time}, the confidence refinement time in EmbodiedOcc++ is only 0.65ms longer than the baseline method, representing a 40.6\% increase in this specific operation. However, this modest increase in refinement time is more than compensated by the 21.5\% reduction in Gaussian updates, which constitutes the primary computational bottleneck in progressive scene reconstruction. Importantly, our approach achieves these efficiency gains without introducing any additional parameters to the model, as the uncertainty estimation leverages the existing semantic feature space. The uncertainty threshold $\tau_{unc}$ is the only hyperparameter added, which is used solely at inference time to determine which points require updates. This parameter-efficient design ensures that our method maintains the same memory footprint as the baseline during training while delivering significant computational savings during inference.

The efficiency gains provided by our method scale with scene complexity and sequence length, with longer sequences benefiting from increasingly selective updates as more regions of the scene reach stability. This scalability is crucial for real-world applications like AR/VR and robotics, where computational resources are often limited and scene reconstruction must occur in real time.

\end{document}